%% file: main.tex
\documentclass[10pt,twocolumn,letterpaper]{article}

\usepackage{cvpr}              %

\input{preamble}

\newcommand{\ourmodel}{\textsc{DiffusionRenderer}\xspace}

\definecolor{cvprblue}{rgb}{0.21,0.49,0.74}
\usepackage[pagebackref,breaklinks,colorlinks,allcolors=cvprblue]{hyperref}

\def\paperID{12055} %
\def\confName{CVPR}
\def\confYear{2025}

\title{\ourmodel: Neural Inverse and Forward Rendering \\with Video Diffusion Models}

\author{
Ruofan Liang$^{1,2,3}$\footnotemark[1], %
Zan Gojcic$^{1}$, 
Huan Ling$^{1,2,3}$, 
Jacob Munkberg$^{1}$,
Jon Hasselgren$^{1}$,
Zhi-Hao Lin$^{1,4}$, \\
Jun Gao$^{1,2,3}$, 
Alexander Keller$^{1}$,
Nandita Vijaykumar$^{2,3}$, 
Sanja Fidler$^{1,2,3}$, 
Zian Wang$^{1,2,3}$\footnotemark[1] \\
$^1$NVIDIA 
\quad $^2$University of Toronto 
\quad $^3$Vector Institute 
\quad $^4$University of Illinois Urbana-Champaign
}

\begin{document}

\twocolumn[{%
\renewcommand\twocolumn[1][]{#1}%
\maketitle
\vspace{-3mm}
\centering
\includegraphics[width=0.95\linewidth]{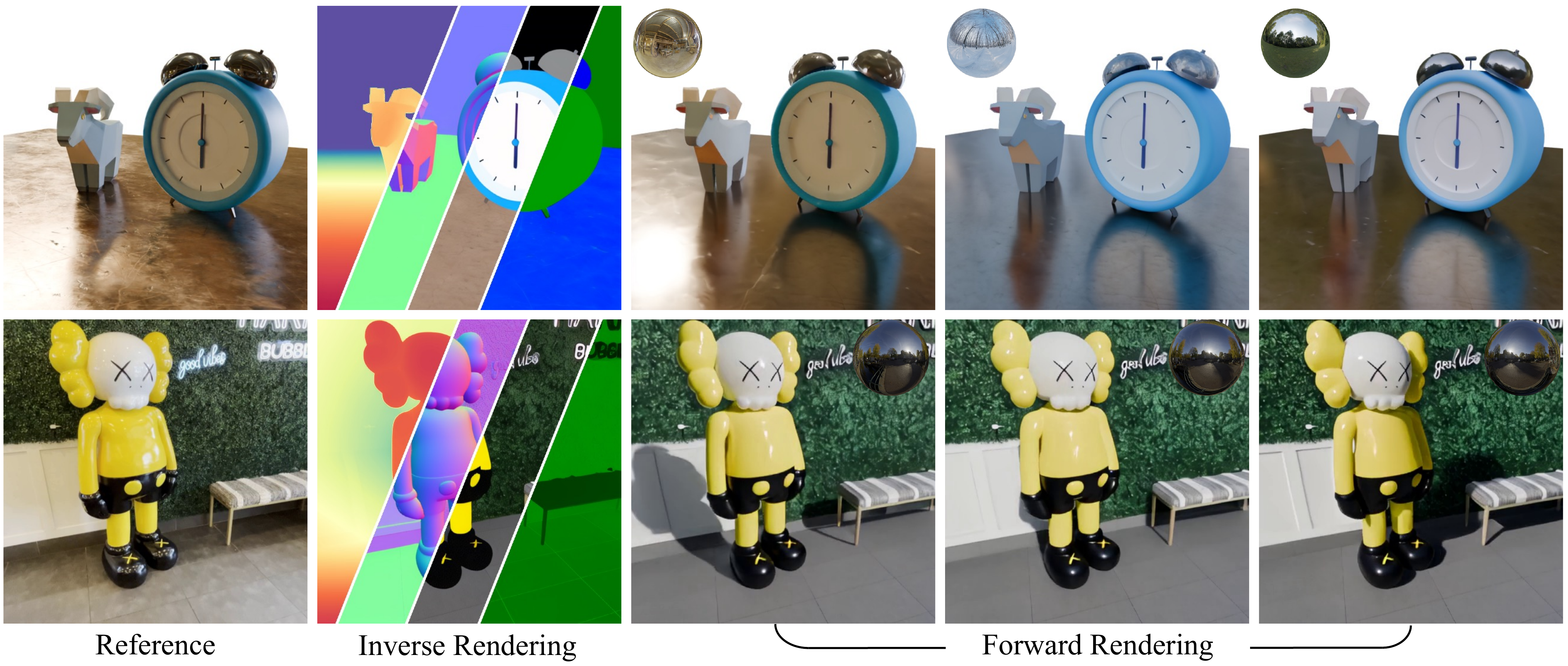}
\vspace{-3mm}
\captionof{figure}{ 
We present \ourmodel{}, a general-purpose method for both neural inverse and forward rendering. 
From input images or videos, it accurately estimates geometry and material buffers, and generates photorealistic images under specified lighting conditions, offering fundamental tools for image editing applications. 
\vspace{1em}
}
\label{fig:teaser}
}]

\begin{abstract}
Understanding and modeling lighting effects are fundamental tasks in computer vision and graphics. Classic physically-based rendering (PBR) accurately simulates the light transport, but relies on precise scene representations--explicit 3D geometry, high-quality material properties, and lighting conditions--that are often impractical to obtain in real-world scenarios. 
Therefore, we introduce \ourmodel{}, a neural approach that addresses the dual problem of inverse and forward rendering within a holistic framework. 
Leveraging powerful video diffusion model priors, the inverse rendering model accurately estimates G-buffers from real-world videos, providing an interface for image editing tasks, and training data for the rendering model.
Conversely,  
our rendering model generates photorealistic images from G-buffers without explicit light transport simulation. 
Specifically, we first train a video diffusion model for inverse rendering on synthetic data, which generalizes well to real-world videos and allows us to auto-label diverse real-world videos. We then co-train our rendering model using both synthetic and auto-labeled real-world data. 
Experiments demonstrate that \ourmodel{} effectively approximates inverse and forwards rendering, consistently outperforming the state-of-the-art. Our model enables practical applications from a single video input—including relighting, material editing, and realistic object insertion. 
\end{abstract}

\input{sec/intro}

\input{sec/related}

\input{sec/method}

\input{sec/results}

\input{sec/conc}

\parahead{Acknowledgments.} 
The authors thank Shiqiu Liu, Yichen Sheng, and Michael Kass for their insightful discussions that contributed to this project.
We also appreciate the discussions with Xuanchi Ren, Tianchang Shen and Zheng Zeng during the model development process.

{
    \small
    \bibliographystyle{ieeenat_fullname}
    \bibliography{main}
}

\input{sec/X_suppl}

\end{document}

%% file: preamble.tex
\usepackage{graphicx}
\usepackage{booktabs}

\usepackage[accsupp]{axessibility}  %

\usepackage[utf8]{inputenc} %
\usepackage[T1]{fontenc}    %
\usepackage{url}            %
\usepackage{booktabs}       %
\usepackage{amsfonts}       %
\usepackage{nicefrac}       %
\usepackage{microtype}      %

\usepackage{caption}
\usepackage{multirow}
\usepackage{xspace}
\usepackage{mathtools}
\usepackage{wasysym}
\usepackage{marvosym}
\usepackage{xcolor}
\usepackage{color}
\usepackage{tabularx}
\usepackage{float}
\usepackage{diagbox,tabu,stackengine}
\usepackage{bbm}
\usepackage{bm}
\usepackage{mwe}
\usepackage{graphbox}
\usepackage{sidecap}

\usepackage[export]{adjustbox} %

\usepackage{algorithm}
\usepackage[noend]{algpseudocode}
\usepackage{comment}
\usepackage{cancel}
\usepackage{ulem}
\usepackage{array}
\newcommand{\PreserveBackslash}[1]{\let\temp=\\#1\let\\=\temp}
\newcolumntype{C}[1]{>{\PreserveBackslash\centering}p{#1}}
\newcolumntype{R}[1]{>{\PreserveBackslash\raggedleft}p{#1}}
\newcolumntype{L}[1]{>{\PreserveBackslash\raggedright}p{#1}}

\usepackage{wrapfig}

\usepackage{xspace}
\makeatletter
\DeclareRobustCommand\onedot{\futurelet\@let@token\@onedot}
\def\@onedot{\ifx\@let@token.\else.\null\fi\xspace}

\def\etal{\textit{et al}\onedot}
\makeatother

\newcommand{\parahead}[1]{\vspace{0.7mm}\noindent\textbf{#1}}

\input{math_commands}

\newcommand{\image}{\mathbf{I}}

\newcommand{\feature}{\mathbf{h}}

\newcommand{\location}{\mathbf{p}}

\newcommand{\normal}{\mathbf{n}}

\newcommand{\light}{L}
\newcommand{\lightdir}{\bm{\omega}}

\newcommand{\brdf}{f_r}

\newcommand{\albedomap}{\mathbf{a}}
\newcommand{\metallicmap}{\mathbf{m}}
\newcommand{\roughnessmap}{\mathbf{r}}
\newcommand{\normalmap}{\mathbf{n}}
\newcommand{\depthmap}{\mathbf{d}}
\newcommand{\envmap}{\mathbf{E}}

\newcommand{\videoLength}{F}
\newcommand{\videoInput}{\mathbf{I}}
\newcommand{\diffusionNoise}{\mathbb{\epsilon}}
\newcommand{\diffusionTime}{\tau}

\newcommand{\diffusionCond}{\mathbf{c}}
\newcommand{\diffusionModel}{\mathbf{f}}
\newcommand{\diffusionModelParams}{\theta}
\newcommand{\videoOutput}{\hat{\mathbf{I}}}
\newcommand{\latent}{\mathbf{z}}

\newcommand{\vaeEncoder}{\mathcal{E}}
\newcommand{\vaeDecoder}{\mathcal{D}}
\newcommand{\dataDistribution}{p_{\text{data}}}
\newcommand{\gbuflatent}{\mathbf{g}}

%% file: math_commands.tex
\usepackage{amsmath,amsfonts,bm}

\def\eqref#1{equation~\ref{#1}}

\def\1{\bm{1}}

\DeclareMathAlphabet{\mathsfit}{\encodingdefault}{\sfdefault}{m}{sl}
\SetMathAlphabet{\mathsfit}{bold}{\encodingdefault}{\sfdefault}{bx}{n}

%% file: sec/intro.tex
\vspace{-3mm}
\section{Introduction}
\label{sec:intro}
\vspace{-1mm}

Understanding and modeling light transport forms the basis of Physically Based Rendering (PBR)~\cite{PBRT}.
Modern path tracing algorithms, as regularly used in the gaming and movie industries,
simulate light transport to render images that cannot be distinguished from photographs.
The quality of such PBR-rendered images heavily depends on the accuracy and realism of the scene's surface geometry, material properties, and lighting representations. Such a scene description is either designed by artists (synthetic scenes) or reconstructed from data---also known as the inverse rendering problem~\cite{Barrow1978RECOVERINGIS,barron2014shape}. Inverse rendering has been extensively studied, particularly for applications like relighting and object insertion into real-world scenes~\cite{li2020inverse,wang2023fegr,zhang2025zerocomp,fortierchouinard2024spotlightshadowguidedobjectrelighting}. However, acquiring high quality surface and material representations is challenging in real-world scenarios, limiting the practicality of PBR methods (Fig.~\ref{fig:rendering_real}).

While physically-based rendering and inverse rendering are usually considered separately, we propose to consider them jointly. 
Our approach draws inspiration from the success of large-scale generative models~\cite{rombach2021highresolution,blattmann2023stable}, which \textit{"render"} photorealistic images from simple text prompts without any explicit understanding of PBR. These models learn the underlying distribution of real-world images from a vast amounts of data, implicitly capturing the complex lighting effects.

Specifically, we propose \ourmodel{}, a general-purpose neural rendering engine that can synthesize light transport simulation---such as shadows and reflections---by leveraging the powerful priors of video diffusion models. Conditioned on input geometry, material buffers, and environment map light source, \ourmodel{} acts as a neural approximation of path-traced shading. \ourmodel{} is designed to remain faithful to the conditioning signals, while adhering to the distribution of real-world images. As a result, we bypass the need for precise scene representations and description, as our model learns to handle imperfections in the input data.

Training such a model requires some amount of high quality and diverse data, including data with noisy conditions to ensure robustness. Therefore, we first train an \textit{inverse renderer}, a video diffusion model to map input RGB videos to intrinsic properties. Although trained solely on synthetic data, the inverse rendering model generalizes robustly to real-world scenarios. We then use it to generate \textit{``pseudo-labels''} for diverse real-world videos. Combining both real-world auto-labeled data and synthetic data, we train our \textit{forward renderer} video diffusion model. 

\ourmodel{} outperforms state-of-the-art methods and effectively approximates the complex functionalities of inverse and forward rendering, 
allowing us to relight images and videos across diverse scenes and to synthesize consistent shadows and reflections without explicit path tracing and 3D scene representation. 
Our model can relight any scene from only a single video input, and provides fundamental tools for editing tasks such as material editing and realistic object insertion. To summarize:
\begin{itemize}[topsep=1pt, leftmargin=10pt]
\item We develop a state-of-the-art inverse rendering method for videos of synthetic and real-world scenes.
\item We repurpose a video diffusion model as a neural rendering engine that can synthesize photorealistic images and videos conditioned on noisy G-buffers. 
\item From a single video input, \ourmodel{} enables relighting, material editing, and virtual object insertion in a unified framework, expanding the possibilities for real-world neural rendering applications. 
\end{itemize}

\input{figures/ssrt}

%% file: figures/ssrt.tex
\begin{figure}[t!]
\centering
{\footnotesize
\setlength{\tabcolsep}{1pt}
\renewcommand{\arraystretch}{1}
\vspace{-4mm}
\begin{tabular}{cccc}
    \raisebox{-0.5\height}{\includegraphics[width=0.24\columnwidth]{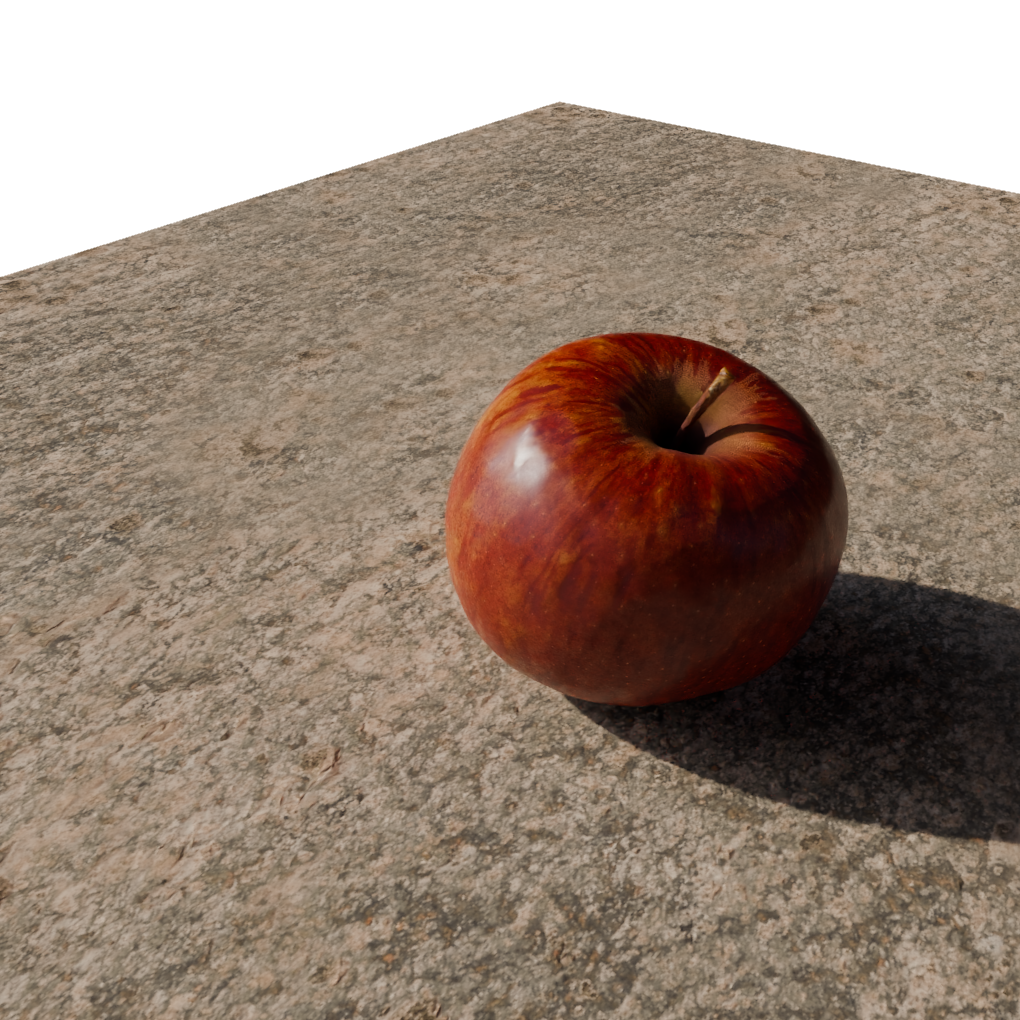}} &
    \raisebox{-0.5\height}{\includegraphics[width=0.24\columnwidth]{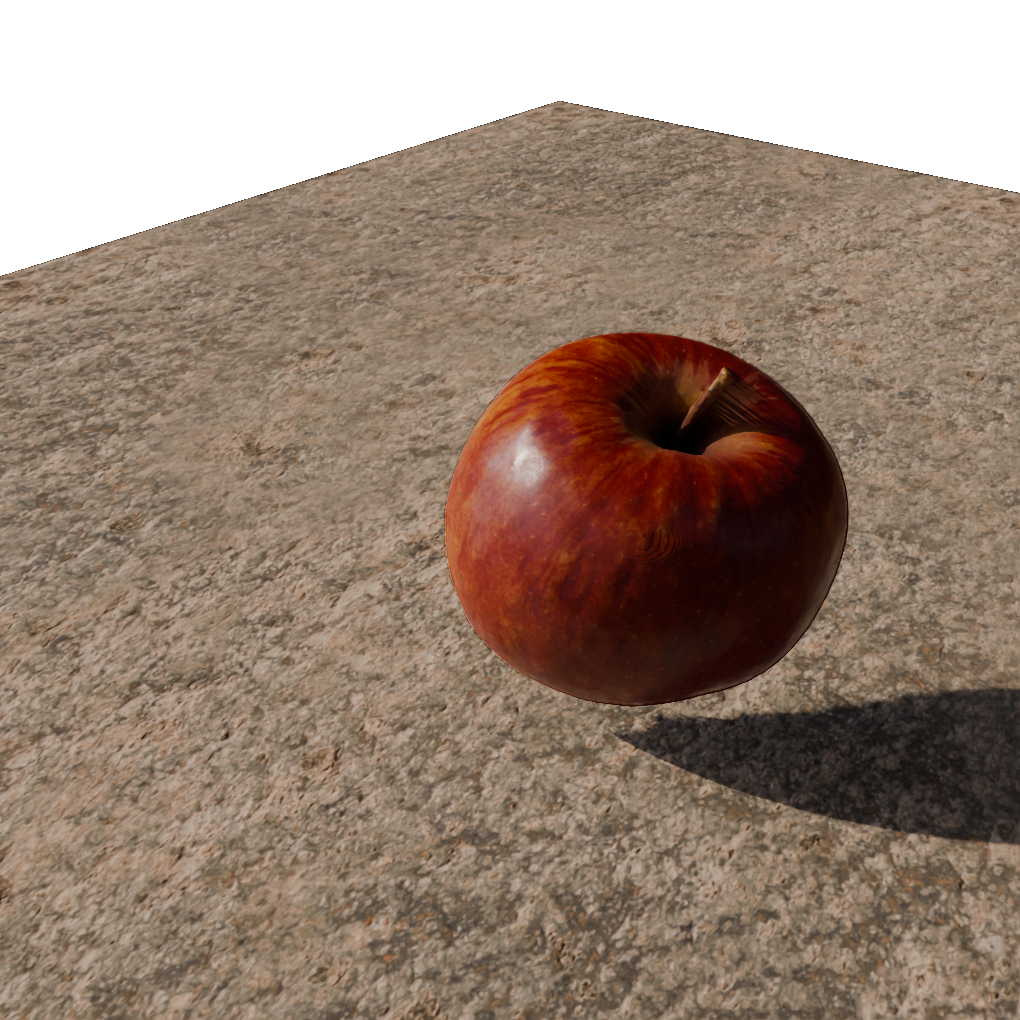}} &
    \raisebox{-0.5\height}{\includegraphics[width=0.24\columnwidth]{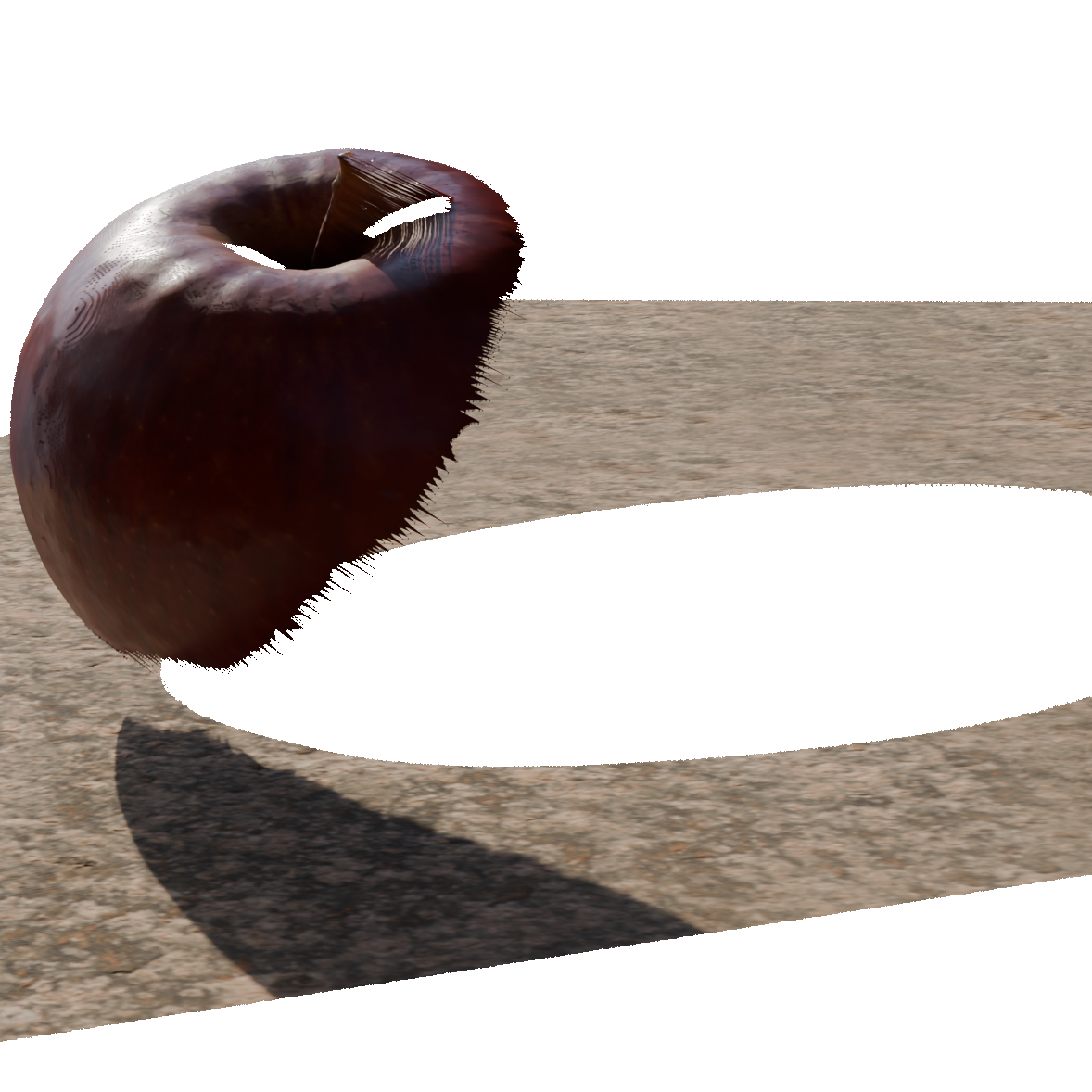}} &
    \raisebox{-0.5\height}{\includegraphics[width=0.24\columnwidth]{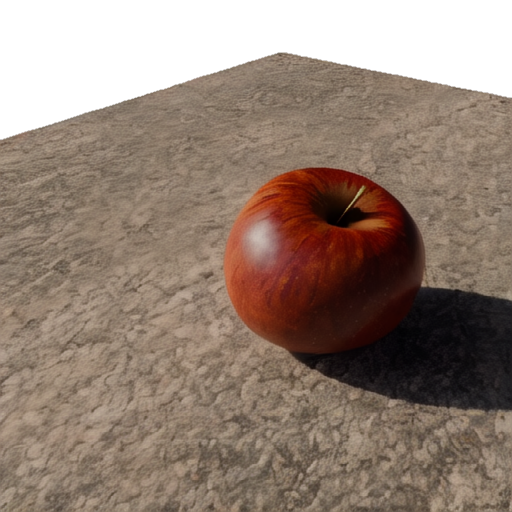}} \\
    Reference & SSRT & SSRT side view & Ours \\
    \raisebox{-0.5\height}{\includegraphics[width=0.24\columnwidth]{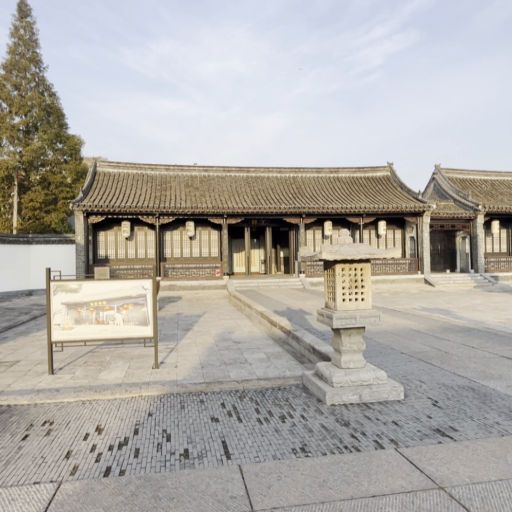}} &
    \raisebox{-0.5\height}{\includegraphics[width=0.24\columnwidth]{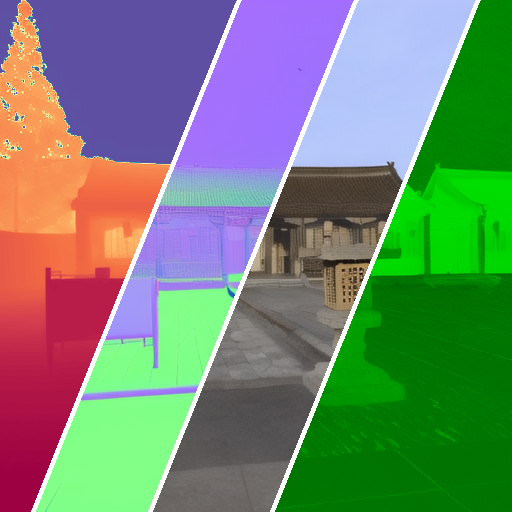}} &
    \raisebox{-0.5\height}{\includegraphics[width=0.24\columnwidth]{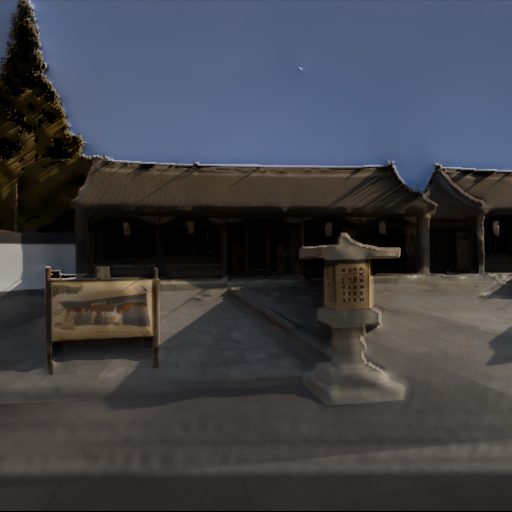}} &
    \raisebox{-0.5\height}{\includegraphics[width=0.24\columnwidth]{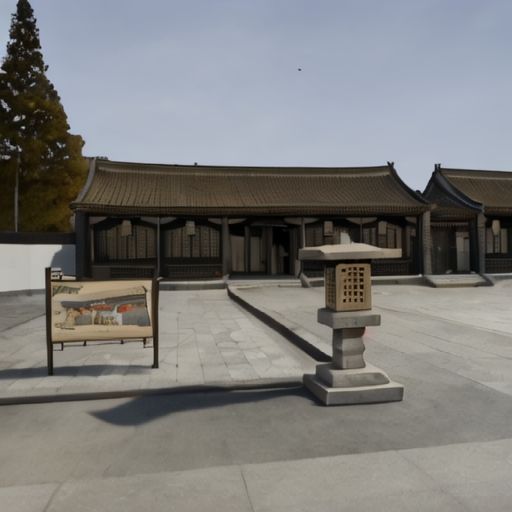}} \\
    Input & Estimated G-buffer & SSRT relit & Ours relit \\
\end{tabular}
}
\vspace{-3mm}
\caption{
Classic PBR relies on explicit 3D geometry, e.g., meshes. When it is not available, screen space ray tracing (SSRT) struggles to accurately represent shadows and reflections (\emph{top}). 
PBR is also sensitive to errors in G-buffers -- SSRT with estimated G-buffers from inverse rendering models often fails to deliver quality results (\emph{bottom}). 
\ourmodel{} bypasses these issues, producing photorealistic results without 3D geometry or perfect G-buffers. 
}
\vspace{-5mm}
\label{fig:rendering_real} 
\end{figure}

%% file: sec/related.tex
\vspace{-3mm}
\section{Related Work}
\vspace{-1mm}

\paragraph{Neural rendering}\hspace{-1.5mm}refers to methods that replace or extend traditional rendering pipelines by neural networks. 
For example, Deep Shading~\cite{nalbach2017deep} replaces traditional deferred 
shading~\cite{deering1988triangle} by a CNN to render images with ambient occlusion, global illumination, and depth-of-field from G-buffers.
More recently, RGB$\leftrightarrow$X~\cite{zeng2024rgb} trains image diffusion models to both estimate a G-buffer from an image and to render an image from a G-buffer. 
We extend this approach to \emph{video} diffusion models and provide a novel approach for neural relighting that does not require an irradiance estimate.
Other approaches fit rendered data using neural models or introduce neural components into an 
existing renderer with focus on approximating light transport~\cite{kallweit2017deep,jiang2021deep} 
or radiance caching~\cite{hermosilla2019deep,muller2021real}. 
A plethora of works on neural and inverse rendering involve volumetric 3D scene representations in the form of NeRF~\cite{mildenhall2020nerf} or 3D Gaussian Splats~\cite{kerbl3Dgaussians}. We refer to~\cite{tewari2020state} for an overview. While providing photo-real view interpolation, these approaches typically bake radiance, and have limited editing capabilities.
In contrast, we explicitly target an intermediate scene representation in the form of traditional, easy-to-edit, G-buffers with separate lighting.

\parahead{Inverse rendering} is a fundamental task first formalized in the 1970s~\cite{barrow1978recovering}, aiming to estimate intrinsic scene properties, like geometry, materials, and lighting from input images. 
Early methods designed hand-crafted priors within an optimization framework~\cite{land1971lightness1,barrow1978recovering,grosse2009ground,bousseau2009user,zhao2012closed,barron2014shape}, typically focusing on low-order effects. These methods lead to errors when the hand-crafted priors do not match reality. 
Recently, supervised and self-supervised learning has been extensively studied~\cite{bell2014intrinsic,kovacs2017shading,li2018cgintrinsics,neuralSengupta19,yu19inverserendernet,li2020inverse,li2020openrooms,Boss2020-TwoShotShapeAndBrdf,wang2021learning,wang2022neural,wimbauer2022rendering,bhattad2023styleganknowsnormaldepth}.
The resulting algorithms are often data-hungry and specific to a certain task or domain. Acquiring sufficient and diverse training data poses a challenge. 
Recent advances in large image generative models provide new deep learning tools for inverse rendering~\cite{intrinsiclora,kocsis2023iid,Phongthawee2023DiffusionLight,liang2024photorealistic,zeng2024rgb}
resulting in much higher reconstruction quality. Still, the quality is not enough to power physically based rendering pipelines.

\parahead{Relighting} focuses on modifying the lighting conditions of a scene given captured images or videos. 
Recent methods reconstruct 3D scene representations from multi-view images, performing explicit inverse rendering to recover material properties and enable relighting~\cite{munkberg2021nvdiffrec,hasselgren2022nvdiffrecmc,chen2021dibrpp,boss2021nerd,physg2021,zhang2021nerfactor,iron-2022,rudnev2022nerfosr,wang2023fegr,liang2023envidr,chen2024intrinsicanything,liang2023gs,shi2023gir,jiang2024gaussianshader}. 
These methods often optimize for each scene individually, and their quality may be affected by practical issues such as single-illumination capture, large scene scale, and dynamic content. 
Learning-based methods that train across multiple scenes have explored latent feature learning~\cite{Zhou_2019_ICCV,liu2020factorize,StyLitGAN} and often incorporate neural rendering modules that utilize PBR buffers as inductive priors~\cite{philip2019multi,griffiths2022outcast,TotalRelighting,kocsis2024lightit,xing2024luminet}. 
To improve relighting quality,
recent approaches~\cite{diffusion_relighting,kocsis2024lightit,zeng2024dilightnet,jin2024neural_gaffer} leverage diffusion models. 
With very few multi-illumination datasets~\cite{murmann19}, existing methods often are specialized to a domain, such as portraits, objects, and outdoor scenes, and remain data-hungry.

%% file: sec/method.tex
\begin{figure*}[t!]
\centering
\vspace{-5mm}
\includegraphics[width=0.95\linewidth]{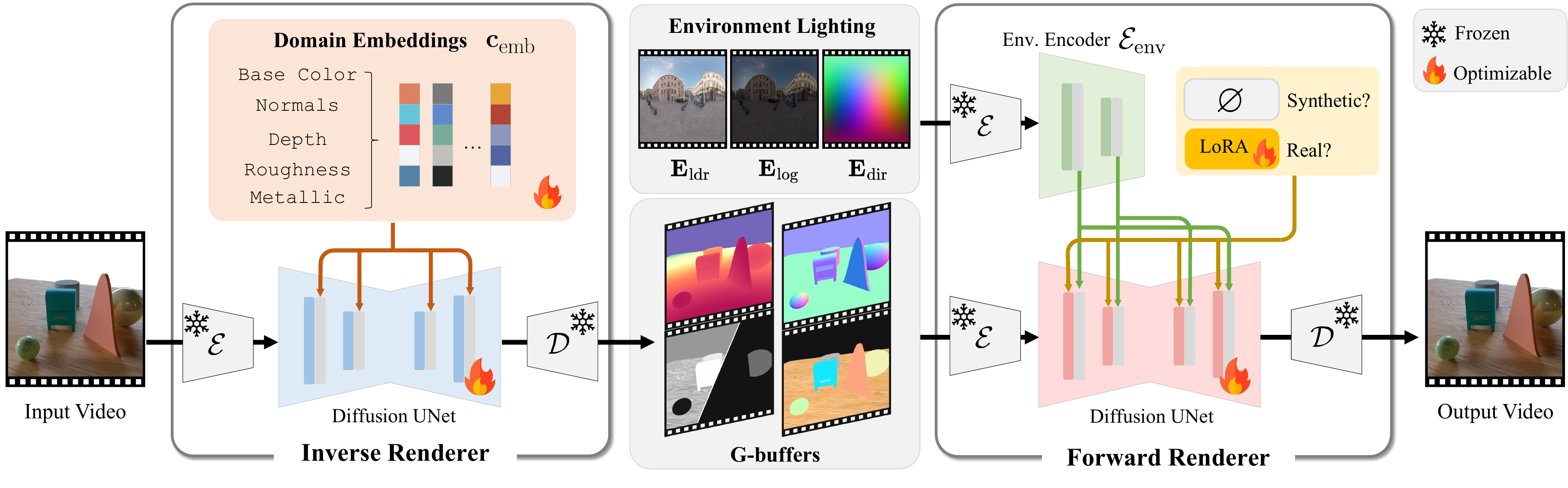}
\vspace{-4mm}
\caption{\textbf{Method overview.} Given an input video, the neural inverse renderer estimates geometry and material properties per pixel. It generates one scene attribute at a time, with the domain embedding indicating the target attributes to generate (Sec.~\ref{sec:neural_inverse_rendering}). 
Conversely, the neural forward renderer produces photorealistic images given lighting information, geometry, and material buffers. The lighting condition is injected into the base video diffusion model through cross-attention layers (Sec.~\ref{sec:neural_rendering}). 
During joint training with both synthetic and real data, we use an optimizable LoRA for real data sources (Sec.~\ref{sec:training}). 
}
\label{fig:method}
\vspace{-4mm}
\end{figure*}

\vspace{-1mm}
\section{Preliminaries} 
\label{sec:preliminaries}
\vspace{-2mm}

\parahead{Physically-based rendering (PBR)}
is concerned with the simulation of how the incoming radiance contributes to the outgoing radiance
\begin{equation}
\light_o(\location, \lightdir_o) = \int_{\Omega} \brdf( \location,  \lightdir_o, \lightdir_i) \light_i( \location,  \lightdir_i) \left | \normal \cdot  \lightdir_i \right | d \lightdir_i,
\label{eq:rendering}
\end{equation}
at a surface point $\location$ in direction $\lightdir_o$.
The integral over the hemisphere $\Omega$ considers the BRDF $\brdf(\location, \lightdir_o, \lightdir_i)$, incoming radiance $\light_i(\location, \lightdir_i)$, and a cosine factor $\left | \normal \cdot \lightdir_i \right |$ for the angle between the normal $\normal$ and incoming light. 
It is evaluated by Monte Carlo methods~\cite{veach1998robust, PBRT}, with meticulously designed BRDF models~\cite{cook1982reflectance,Walter2007,burley2012physically} that approximate real surfaces.

\parahead{Video diffusion models (VDMs).} 
A diffusion model learns to approximate a data distribution $\dataDistribution(\videoInput)$ via iterative denoising~\cite{sohl2015deep, ho2020denoising, dhariwal2021diffusion}. 
Most VDMs operate in a compressed, lower-dimensional latent space~\cite{blattmann2023stable,cosmos}. 
Given an RGB video $\videoInput\in\mathbb{R}^{\videoLength \times H \times W \times 3}$, consisting of $\videoLength$ frames at resolution $H \times W$, a pre-trained VAE encoder $\vaeEncoder$ first encodes the video into a latent representation $\latent = \vaeEncoder(\videoInput) \in\mathbb{R}^{\videoLength^\prime \times h \times w \times C }$. The final video $\videoOutput$ is then reconstructed by decoding $\latent$ with a pre-trained VAE decoder $\vaeDecoder$. Both training and inference stages of the VDM are conducted in this latent space.
In this work, we build on Stable Video Diffusion~\cite{blattmann2023stable}, which compresses the video only along the spatial dimensions: $\videoLength^{\prime} = \videoLength$, $C = 4$, $h = \frac{H}{8}$, and $w = \frac{W}{8}$. 

To train the VDM, noisy versions $\latent_\diffusionTime = \alpha_\diffusionTime \latent_0 + \sigma_\diffusionTime \diffusionNoise$ are constructed by adding a Gaussian noise $\diffusionNoise$
with the noise schedule provided by $\alpha_\diffusionTime$ and $\sigma_\diffusionTime$ following EDM~\cite{Karras2022edm}. 
The diffusion model parameters $\diffusionModelParams$ of the denoising function $\diffusionModel_{\diffusionModelParams}$ are optimized using the denoising score matching objective~\cite{Karras2022edm}. 
Once trained, iteratively applying $\diffusionModel_{\diffusionModelParams}$ to a sample of Gaussian noise will produce a sample of $\dataDistribution(\videoInput)$.

\parahead{Conditioning in VDMs.} 
Two common approaches to inject conditions into VDMs are: (i) concatenating condition channels with image latents $\latent_\diffusionTime$, which is often used for pixel-wise conditions~\cite{blattmann2023stable,ke2023repurposing,kocsis2023iid,zeng2024rgb}, and (ii) injecting conditions through cross-attention layers~\cite{rombach2021highresolution,blattmann2023stable}, which is often used for semantic features such as the CLIP embedding~\cite{CLIP}. %
Note that our method is compatible with any standard VDMs and does not depend on specific architectural details.

\vspace{-1mm}
\section{Method} 
\label{sec:method}
\vspace{-1mm}
\ourmodel{} is a unified framework comprising two video diffusion models designed for the dual tasks of neural forward and inverse rendering. As illustrated in Fig.~\ref{fig:method},
the \textit{neural forward renderer} (Sec.~\ref{sec:neural_rendering}) approximates physically based light transport (Eq.~\ref{eq:rendering}), 
transforming G-buffers \cite{nalbach2017deep} and lighting into a photorealistic video.
The \textit{neural inverse renderer} (Sec.~\ref{sec:neural_inverse_rendering}) reconstructs geometry and material buffers from input video. 
The neural forward and inverse renderers are based on pre-trained video diffusion models and fine-tuned for conditional generation~\cite{ke2023repurposing,kocsis2023iid,zeng2024rgb}. 

Data is a critical aspect of learning-based methods. We describe our data curation workflow and synthetic-real joint training strategies in Sec.~\ref{sec:data} and Sec.~\ref{sec:training}. 
Finally, we discuss image editing applications in Sec.~\ref{sec:editing_application}.

\subsection{Neural Forward Rendering}
\label{sec:neural_rendering}
\vspace{-1mm}
We formulate neural forward rendering as a conditional generation task, producing photorealistic images given geometry, materials, and lighting as conditions.
By approximating light transport simulation in a data-driven manner, the model requires neither classic 3D geometry nor explicit path tracing, thus reducing the constraints in real-world applications. 

\parahead{Geometry and material conditions.}
Similar to the G-buffers in rendering system based
on deferred shading~\cite{deering1988triangle}, we use per-pixel scene attribute maps to represent scene geometry and materials. 
Specifically, we use surface normals $\normalmap \in \mathbb{R}^{\videoLength \times H\times W\times 3}$ in camera space and relative depth $\depthmap \in \mathbb{R}^{\videoLength \times H\times W\times 1}$ normalized to $[-1,1]$ to represent scene geometry. %
For materials, we use base color $\albedomap \in \mathbb{R}^{\videoLength \times H\times W\times 3}$, roughness $\roughnessmap \in \mathbb{R}^{\videoLength \times H\times W\times 1}$, and metallic $\metallicmap \in \mathbb{R}^{\videoLength \times H\times W\times 1}$ following the Disney BRDF~\cite{burley2012physically}. 

\parahead{Lighting conditions.} 
Lighting is represented by environment maps $\envmap \in \mathbb{R}^{\videoLength \times H_\text{env}\times W_\text{env}\times 3}$, which are panoramic images that capture the lighting intensity from all directions over the sphere. 
These environment maps are encoded in high dynamic range (HDR), while the VAEs used in typical latent diffusion models are designed for pixel values between $-1$ and $1$.
To address this discrepancy, similar to the light representation in Neural Gaffer~\cite{jin2024neural_gaffer}, we first apply Reinhard tonemapping to convert HDR environment map into an LDR image $\envmap_\text{ldr}$. 
To more effectively represent HDR values, particularly for light sources with high-intensity peaks, we compute $\envmap_\text{log} = \log (\envmap + 1) / E_\text{max}$ where the light intensity values are mapped to logarithm space that is closer to human perception and normalized by max log intensity $E_\text{max}$. 
Additionally, we also compute a directional encoding image, $\envmap_\text{dir} \in \mathbb{R}^{\videoLength \times H_\text{env}\times W_\text{env}\times 3}$, where each pixel is represented by a unit vector indicating its direction in the camera coordinate system. 
The resulting lighting encodings used by the model consist of three panoramic images: $\{ \envmap_\text{ldr}, \envmap_\text{log}, \envmap_\text{dir} \}$.

\parahead{Model architecture.} 
Our models are based on Stable Video Diffusion~\cite{blattmann2023stable}, an image-to-video diffusion model with its core architecture including a VAE encoder-decoder pair $\{\vaeEncoder, \vaeDecoder\}$, and a UNet-based denoising function $\diffusionModel_\diffusionModelParams$.

We use the VAE encoder $\vaeEncoder$ to separately encode each G-buffer from $\{\normalmap, \depthmap, \albedomap, \roughnessmap, \metallicmap\}$  into the latent space and concatenate them to produce the pixel-aligned scene attribute latent map $\gbuflatent  = \{\vaeEncoder(\normalmap),\vaeEncoder(\depthmap),\vaeEncoder(\albedomap),\vaeEncoder(\roughnessmap),\vaeEncoder(\metallicmap)\} \in \mathbb{R}^{\videoLength \times h \times w \times 20}$.

Environment maps are usually in equi-rectangular projection and are not pixel-aligned with the generated images, thus requiring extra consideration. 
Prior works explored directly concatenating environment maps to the image latents~\cite{jin2024neural_gaffer} or concatenate split-sum shading buffers~\cite{deng2024flashtex}, which we also experimented with, but found suboptimal (Table~\ref{tab:rendering}). 
Instead, we take the cross-attention layers which originally operate on the text/image CLIP features, and re-purpose them for lighting conditions. To preserve spatial details of the environment maps, we generalize the conditional signals to a list of multi-resolution feature maps. 

Specifically, we first pass the environment map information through VAE encoder $\vaeEncoder$ to obtain $\feature_\envmap = \{ \vaeEncoder(\envmap_\text{ldr}), \vaeEncoder(\envmap_\text{log}), \vaeEncoder(\envmap_\text{dir}) \} \in \mathbb{R}^{\videoLength \times h_\text{env} \times w_\text{env} \times 12}$. 
We additionally use an environment map encoder $\vaeEncoder_\text{env}$ to further operate on $\feature_\envmap$. 
$\vaeEncoder_\text{env}$ is the simplified encoder part of diffusion UNet with attention and temporal layers removed. It contains several convolutional layers to downsample and extract $K$ levels of multi-resolution features as lighting conditions: 
\begin{equation}
\diffusionCond_\text{env} := \{ \feature_\text{env}^i \}_{i=1}^K = \vaeEncoder_\text{env}(\feature_\envmap) 
\label{eq:lighting_cond}
\end{equation}

As a result, the diffusion UNet $\diffusionModel_{\diffusionModelParams}$ takes the noisy latent $\latent_\diffusionTime$ and G-buffer latent $\gbuflatent$ as pixel-wise input. At each UNet level $k$, the diffusion UNet \textit{queries} the latent environment map features at the corresponding level $\feature_\text{env}^k$, and aggregates based on its \textit{keys} and \textit{values}. 
Through the multi-level self-attention and cross-attention layers, the diffusion model is given the capacity to learn to shade G-buffers with lighting. 
During inference, the diffusion target can be computed as $\diffusionModel_{\diffusionModelParams}(\latent_\diffusionTime; \gbuflatent, \diffusionCond_\text{env}, \diffusionTime)$ to produce photorealistic images with iterative denoising.

\subsection{Neural Inverse Rendering}
\label{sec:neural_inverse_rendering} 
\vspace{-1mm}

We similarly formulate inverse rendering as a conditional generation task. 
Given an input video $\videoInput$ as a condition, the inverse renderer estimates scene attribute maps 
$\{ \normalmap, \depthmap, \albedomap, \roughnessmap, \metallicmap\}$ which are the G-buffers used by the forward renderer.

\parahead{Model architecture.} 
The input video $\videoInput$ is encoded into latent space $\latent = \vaeEncoder(\videoInput)$, and concatenated with the noisy G-buffer latent, which we denote as $\gbuflatent_\diffusionTime= \alpha_\diffusionTime \gbuflatent_0 + \sigma_\diffusionTime \diffusionNoise$. 

Given an input video, the inverse renderer generates all five attributes $\{\normalmap, \depthmap, \albedomap, \roughnessmap, \metallicmap\}$ using the same model. 
To preserve the high-quality generation and maximally leverage the diffusion model pre-trained knowledge, each attribute is generated in a dedicated pass, instead of generating all at once. We follow prior works~\cite{long2023wonder3d,fu2024geowizard,zeng2024rgb} and use a domain embedding to indicate to the model which attribute should be generated. 
Specifically, we introduce an optimizable domain embedding $\diffusionCond_\text{emb} \in \mathbb{R}^{K_\text{emb} \times C_\text{emb}}$, where $K_\text{emb} = 5$ is the number of buffers and $C_\text{emb}$ is the dimension of the embedding vector.
We re-purpose the cross-attention layers with image CLIP features to take domain embeddings. 
When estimating an attribute indexed by $P$, 
we feed its embedding $\diffusionCond_\text{emb}^P$ as a condition and predict the diffusion target with $\diffusionModel_{\diffusionModelParams}(\gbuflatent^P_\diffusionTime; \latent, \diffusionCond_\text{emb}^P, \diffusionTime)$.

\subsection{Data Strategy} 
\label{sec:data}
\vspace{-1mm}
\parahead{Synthetic data curation.} 
To train our models, we require high-quality video data with paired ground-truth for material, geometry, and lighting information. 
Specifically, each video data sample should include paired frames of RGB, base color, roughness, metallic, normals, depth, and environment map: $\{\videoInput, \albedomap, \roughnessmap, \metallicmap, \normalmap, \depthmap, \envmap \}$.  
These buffers are typically only available in synthetic data, and most existing public datasets contain only a subset of them.

To address the data scarcity, we designed a synthetic data generation workflow to produce a large amount of high-quality data covering diverse and complex lighting effects.  We start by curating a collection of 3D assets, PBR materials, and HDRI environment maps. 
We use 36,500 3D assets from Objaverse LVIS split. 
For materials and lighting, we collected 4,260 high-quality PBR material maps, and 766 HDR environment maps from publicly available resources. 

In each scene, we place a plane with a randomly selected PBR material, and sample up to three 3D objects, and place them on the plane. We perform collision detection to avoid intersecting objects. 
We also place up to three primitives (cube, sphere, and cylinder) with randomized shape and materials to cover complex lighting effects such as inter-reflections. A randomly selected HDR environment map illuminates the scene. 
We generate motions including camera orbits, camera oscillation, lighting rotation, object rotation and translation. 

We use a custom path tracer based on OptiX~\cite{parker2010optix} to render the videos. 
In total, we generate 150,000 videos with paired ground-truth G-buffers and environment maps, at 24 frames per video in 512x512 resolution. This dataset can be used to train both rendering and inverse rendering models.

\parahead{Real world auto-labeling.} 
Synthetic data provides accurate supervision signals, and when combined with powerful image diffusion models, it demonstrates impressive generalization to unseen domains for inverse rendering tasks~\cite{ke2023repurposing,fu2024geowizard}. 
However, when it comes to training the forward rendering model, synthetic data alone is insufficient. Since the output of the forward renderer is an RGB video, training only on synthetic renderings biases the model toward synthetic visual styles. 
Compared to inverse rendering, we observe a much more significant domain gap in complex real-world scenes for forward rendering tasks (Fig.~\ref{fig:relighting_ablation}).

Acquiring real-world data with paired geometry, material, and lighting ground truth requires complex and impractical capturing setups. Based on the observation that our inverse rendering model generalizes to real-world videos, we apply it to automatically label real-world videos. 
Specifically, we use the DL3DV10k~\cite{ling2024dl3dv} dataset, which is a large-scale dataset consisting of 10,510 videos featuring diverse real-world environments. 
We use our inverse rendering model (Sec.~\ref{sec:neural_inverse_rendering}) to generate G-buffer labels and use an off-the-shelf method DiffusionLight~\cite{Phongthawee2023DiffusionLight} to estimate environment maps. 
Each video is divided into 15 segments, resulting in around 150,000 real-world video samples with auto-labeled geometry, material, and lighting attributes.

\subsection{Training pipeline} 
\label{sec:training}
\vspace{-1mm}

\parahead{Neural inverse renderer.} 
We first co-train the inverse rendering model on the combination of the curated synthetic video dataset and public image intrinsics datasets InteriorVerse~\cite{zhu2022learning} and HyperSim~\cite{Hypersim}. For image datasets, we treat images as single-frame videos. Each data sample consists of a video $\videoInput$, an attribute index $P$, and the scene attribute map $\mathbf{s}^P$. 
The target latent variable is the latents of the scene attribute $\gbuflatent^P_0 := \vaeEncoder(\mathbf{s}^P)$, and noise is added to $\gbuflatent^P_0$ to produce $\gbuflatent^P_\diffusionTime$. The model is trained using the objective function~\cite{Karras2022edm}:
\begin{equation}
\mathcal{L}(\diffusionModelParams, \diffusionCond_\text{emb}) = \|\diffusionModel_{\diffusionModelParams} \left(\gbuflatent^P_\diffusionTime; \latent, \diffusionCond_\text{emb}^P, \diffusionTime \right) - \gbuflatent^P_0\|_2^2.
\end{equation}
We fine-tune the diffusion model parameters $\diffusionModelParams$ and domain embeddings $\diffusionCond_\text{emb}$, while keeping the latent encoder $\vaeEncoder$ and decoder $\vaeDecoder$ frozen. Once trained, the inverse renderer is used to auto-label real-world videos, generating training data for the forward renderer.

\parahead{Environment map encoder pre-training.} 
Following the approach of latent diffusion models~\cite{rombach2021highresolution}, we pre-train the environment map encoder $\vaeEncoder_\text{env}$ along with a decoder $\vaeDecoder_\text{env}$ using an L2 image reconstruction objective on environment maps, similar to an auto-encoder. The decoder architecture is based on the UNet decoder, containing a set of upsampling layers. 
After training, we discard the decoder $\vaeDecoder_\text{env}$ and freeze the environment map encoder $\vaeEncoder_\text{env}$ while training the neural forward rendering model.

\parahead{Neural forward renderer.} 
We train our rendering model on a combination of a synthetic video dataset and real-world auto-labeled data, using paired G-buffer, lighting, and RGB videos.
Although the auto-labeled real-world data is of sufficient quality, it may still contain inaccuracies. 
To address discrepancies between the synthetic and real-world data sources, we introduce an additional LoRA~\cite{hu2022lora} with a small set of optimizable parameters $\Delta\diffusionModelParams$ during training on real data. We empirically find it improves the rendering quality (Fig.~\ref{fig:relighting_ablation}). 

During training, for an RGB video $\mathbf{I}$, the target latent variable is defined as $\latent_0 := \vaeEncoder(\mathbf{I})$. Noise is added to $\latent_0$ to produce noisy image latent $\latent_\diffusionTime$. The training objective is:
\vspace{-6mm}
\begin{align}
\mathcal{L} (\diffusionModelParams, \Delta\diffusionModelParams) =& \|\diffusionModel_{\diffusionModelParams} \left( \latent^\texttt{synth}_\diffusionTime; \gbuflatent^\texttt{synth}, \diffusionCond^\texttt{synth}_\text{env}, \diffusionTime \right) -  \latent^\texttt{synth}_0\|_2^2 \;+ \notag \\ 
& \|\diffusionModel_{\diffusionModelParams + \Delta\diffusionModelParams}\left(\latent_\diffusionTime^\texttt{real}; \gbuflatent^\texttt{real}, \diffusionCond_\text{env}^\texttt{real}, \diffusionTime \right) -  \latent_0^\texttt{real} \|_2^2  
\end{align}

\subsection{Editing Applications} 
\label{sec:editing_application}
\vspace{-1mm}
Our proposed framework provides fundamental solutions for inverse and forward rendering, enabling photorealistic image editing applications through a three-step process: neural inverse rendering, G-buffer and lighting editing, and neural rendering. 
For the example of relighting, given a video $\videoInput$ as input, the inverse rendering model estimates the G-buffers 
\begin{equation}
\{ \hat{\normalmap}, \hat{\depthmap}, \hat{\albedomap}, \hat{\roughnessmap}, \hat{\metallicmap} \} = \texttt{InverseRenderer}(\videoInput).
\end{equation}
With a user-specified target environment map $\envmap_\text{tgt}$, the rendering model produces relit videos 
\begin{equation}
\hat{\videoInput}_\text{tgt}=\texttt{ForwardRenderer}(\{ \hat{\normalmap}, \hat{\depthmap}, \hat{\albedomap}, \hat{\roughnessmap}, \hat{\metallicmap},  \envmap_\text{tgt} \}).
\end{equation}
Similarly, editing the G-buffers and rendering the videos can enable material editing and virtual object insertion.

%% file: sec/results.tex
\input{tables/rendering}

\input{tables/relighting}

\input{tables/inverse_rendering_combined}

\section{Experiments}
\label{sec:experiments}
\vspace{-1mm}

\begin{figure*}[t!]
\centering
\includegraphics[width=0.98\linewidth]{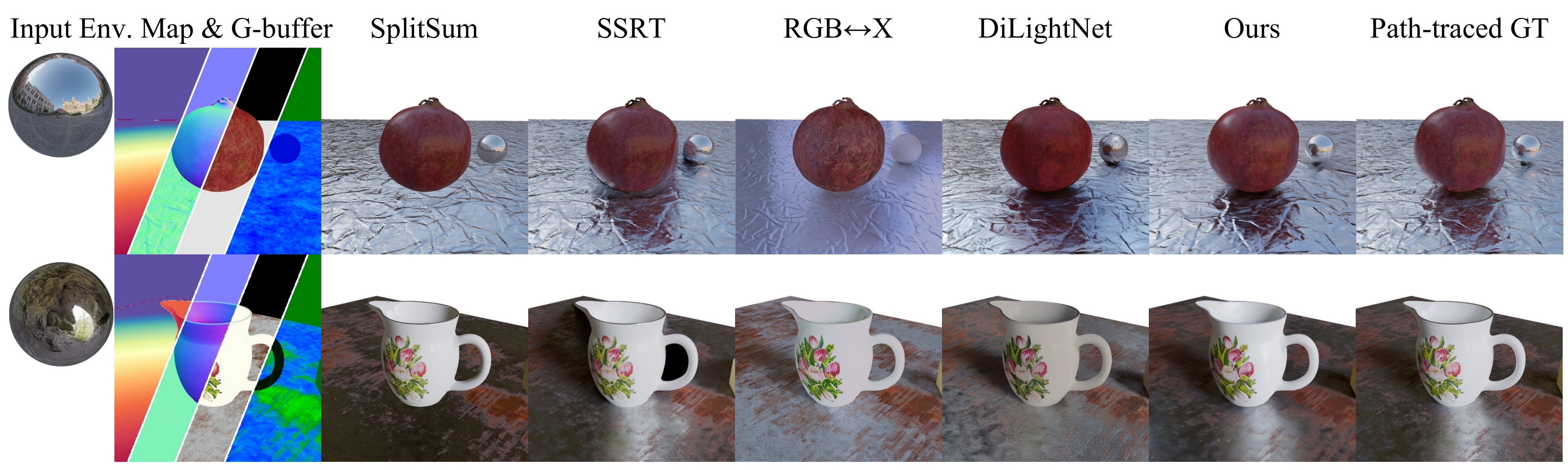}
\vspace{-3mm}
\caption{
\textbf{Qualitative comparison of forward rendering.} 
Our method generates high-quality inter-reflections (\textit{top}) and shadows (\textit{bottom}), producing more accurate results than the neural baselines. 
}
\vspace{-3.5mm}
\label{fig:rendering}
\end{figure*}

\begin{figure*}[t!]
\centering
\includegraphics[width=0.98\textwidth]{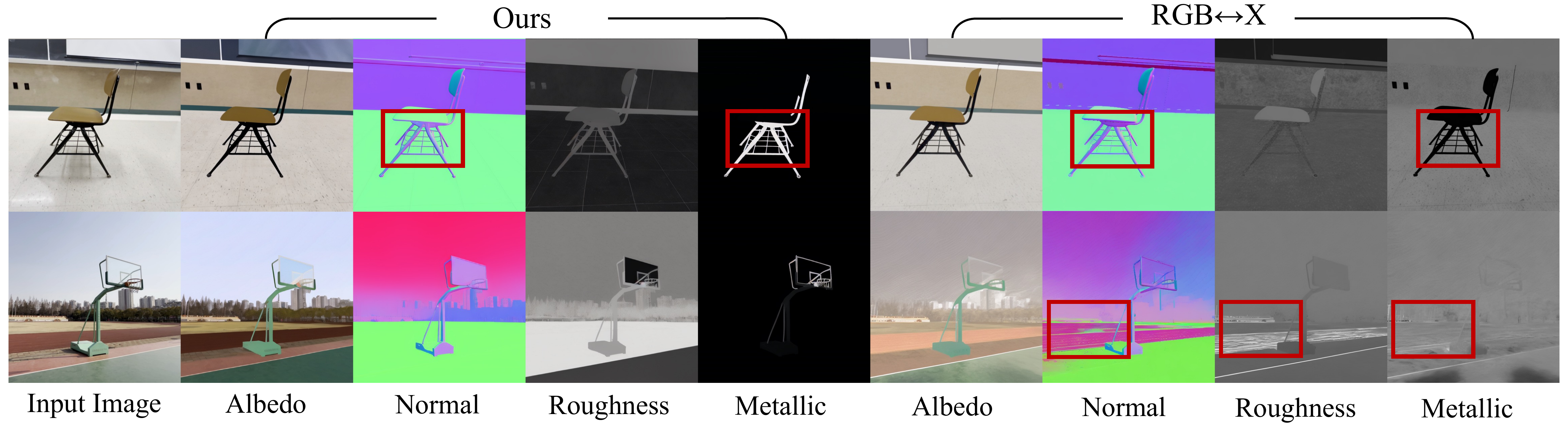}
\vspace{-3mm}
\caption{
\textbf{Qualitative comparison of inverse rendering.} We compare with RGB$\leftrightarrow$X~\cite{zeng2024rgb} on DL3DV10k dataset. Both methods work well on indoor scenes, while our method predicts finer details in thin structures and more accurate metallic and roughness channels (top), likely benefiting from our curated training data. 
As compared to RGB$\leftrightarrow$X, our method generalizes better to outdoor scenes (bottom row). 
}
\vspace{-4.5mm}
\label{fig:inverse_rendering_real}
\end{figure*}

We extensively evaluate \ourmodel{} on a diverse range of synthetic and real-world datasets. 
Sec.~\ref{sec:exp_settings} details our experimental settings.
We compare and ablate across three main tasks: image generation from G-buffers (Sec.~\ref{sec:exp_rendering}), inverse rendering (Sec.~\ref{sec:exp_inverse_rendering}), and relighting (Sec.~\ref{sec:exp_relighting}).  
Finally, we show applications of our method in Sec.~\ref{sec:exp_applications}.

\subsection{Experiment Settings}
\label{sec:exp_settings} 
\vspace{-1mm}
We refer to model implementation details in the Supplement. 

\parahead{Task definitions.} 
We evaluate our method on three fundamental tasks: forward rendering, inverse rendering, and relighting.
For forward rendering, the methods take the G-buffers and lighting information $\{\normalmap, \depthmap,  \albedomap, \roughnessmap, \metallicmap, \envmap \}$ as input, and output rendered images $\hat{\videoInput}$. We evaluate the consistency between the rendered outputs and the ground-truth images $\videoInput$. 

For inverse rendering, each method uses RGB images $\videoInput$ as input to estimate scene attributes $\{ \hat{\albedomap}, \hat{\roughnessmap}, \hat{\metallicmap}, \hat{\normalmap} \}$, and compare against ground truth values. 
Our focus is primarily on the attributes related to PBR -- specifically, base color, roughness, and metallic properties. We recognize dedicated works on normal and depth estimation~\cite{ke2023repurposing,fu2024geowizard} and do not aim to provide an exhaustive evaluation. 
For relighting, each method takes RGB images and target lighting conditions $\{ \videoInput^\text{src}, \envmap^\text{tgt} \}$ as input, output re-lit image sequence $\hat{\videoInput}^\text{tgt}$ under the target lighting conditions, and compare with ground truth $\videoInput^\text{tgt}$.

\parahead{Baselines.} 
For forward rendering, we compare with Split Sum~\cite{karis2013real} and Screen Space Ray Tracing (SSRT). 
For SSRT, we extract a mesh from the depth buffer and render the mesh with material parameters from the G-buffers and a provided HDR probe in a path tracer.
We additionally compare against the neural rendering components of recent state-of-the-art methods RGB$\leftrightarrow$X~\cite{zeng2024rgb} and DiLightNet~\cite{zeng2024dilightnet}. 
For inverse rendering, we compare with recent diffusion-based methods Kocsis~\etal~\cite{kocsis2023iid}, RGB$\leftrightarrow$X~\cite{zeng2024rgb} and earlier methods~\cite{li2020inverse,zhu2022learning,bell2014intrinsic}. 
For relighting, we compare with 2D methods DiLightNet~\cite{zeng2024dilightnet}, Neural Gaffer~\cite{jin2024neural_gaffer}.
We also compare with 3D reconstruction-based methods~\cite{wang2023fegr,lin2023urbanir} in supplement.

\parahead{Metrics.} 
We use PSNR, SSIM, and LPIPS~\cite{zhang2018unreasonable} for forward rendering and relighting. 
For inverse rendering, we evaluate albedo with PSNR and LPIPS following~\cite{kocsis2023iid,zeng2024rgb}. 
Since albedo estimation involves scale ambiguity~\cite{grosse2009ground}, we additionally solve and apply a three-channel scaling factor using least-squares error minimization before computing metrics, referred to as si-PSNR and si-LPIPS. %
We use root mean square error (RMSE) for metallic and roughness evaluation, and mean angular error for normals. 

\parahead{Datasets.} 
We curate two high-quality synthetic datasets for quantitative evaluation, named \textit{SyntheticScenes} and \textit{SyntheticObjects}. The datasets consist of 3D assets from PolyHaven~\cite{polyhaven} and Objaverse~\cite{objaverse} that are not included in the training data of our method or the baseline methods.
\textit{SyntheticScenes} contains 40 scenes, each featuring multiple objects arranged on a plane textured with high-quality PBR materials. Each scene is rendered into 24-frame videos under four lighting conditions, with motions such as camera orbiting and oscillation.
As some baseline methods perform best with object-centric setups, we also create \textit{SyntheticObjects}, a dataset of 30 individual objects. 
For each object, we render 24-frame videos under four different lighting conditions, with lighting rotated across frames.

For inverse rendering, we also evaluate on the indoor scene benchmark InteriorVerse~\cite{zhu2022learning}. We include qualitative comparisons on the DL3DV10k~\cite{ling2024dl3dv} dataset.

\subsection{Evaluation of Forward Rendering}
\label{sec:exp_rendering}
\vspace{-1mm}

We compare our method with baseline methods in Table~\ref{tab:rendering} and Fig.~\ref{fig:rendering}. 
For the neural methods RGB$\leftrightarrow$X~\cite{zeng2024rgb} and DiLightNet~\cite{zeng2024dilightnet}, we use their rendering models with ground-truth G-buffers to generate the images.

Both classic PBR and neural methods perform well on the \textit{SyntheticObjects} dataset in single-object settings but show significant quality drops on the \textit{SyntheticScenes} dataset due to complex inter-reflections and occlusions. 
For example, our method exhibits a minor PSNR decrease of $2.3$~dB from \textit{SyntheticObjects} to \textit{SyntheticScenes}, while other baselines show more substantial drops.

Our method consistently outperforms all neural methods on both datasets and performs comparably to classic methods. 
In real-world editing applications, however, these PBR techniques often face significant limitations due to missing 3D geometry and noisy G-buffers (Fig.~\ref{fig:rendering_real}). Furthermore, SplitSum does not model shadows and inter-reflections.

\parahead{Ablation study.} 
We ablate our model design choices in Table~\ref{tab:rendering}. 
While our method generalizes from images to videos, it can treat images as a special case of single-frame videos. To evaluate the benefits introduced by the video model, we compare with an ablated version of our method that performs per-frame inference (Ours, \texttt{image}). The video model consistently improves rendering quality across both datasets. 
In the ablated variant Ours (w/o env.~encoder), we concatenate the VAE encoded environment maps directly to the image channels~\cite{jin2024neural_gaffer} rather than using a separate environment map encoder. We demonstrate that a dedicated environment map encoder improves performance.
For Ours (+ shading cond.), we include split-sum shading buffers as a conditioning input, following~\cite{deng2024flashtex}. However, we observed no significant improvement with this addition and chose to exclude it from our final method for simplicity.

\vspace{-1mm}
\subsection{Evaluation of Inverse Rendering} 
\label{sec:exp_inverse_rendering}
\vspace{-1mm}
We quantitatively compare our method with baseline methods on \textit{SyntheticScenes} in Table~\ref{tab:inverse_rendering} and InteriorVerse~\cite{zhu2022learning} benchmark in Table~\ref{tab:inverse_rendering_indoor}. Our methods consistently outperforms baseline methods in both datasets, indicating the effectiveness of our data curation workflow and method designs. 
We show a qualitative comparison with RGB$\leftrightarrow$X~\cite{zeng2024rgb} on DL3DV10k~\cite{ling2024dl3dv} dataset in Fig.~\ref{fig:inverse_rendering_real}.

\parahead{Image vs.~video model.} 
Comparing Ours (\texttt{image}) and Ours, the video model consistently enhances the quality of inverse rendering across all attributes.
Notably, for properties associated with specular materials, the video model reduces RMSE by $41\%$ for metallic (from 0.066 to 0.039) and $20\%$ for roughness (from 0.098 to 0.078) compared to the image model. This suggests that the model learns to leverage view changes in video data, effectively capturing view-dependent effects to predict specular properties more accurately.

\parahead{One-step deterministic fine-tuning.} 
We also ablate the design choice of 1-step deterministic fine-tuning in Table~\ref{tab:inverse_rendering} and~\ref{tab:inverse_rendering_indoor}. 
By default, the inverse renderer performs 20 denoising steps during inference.
Building on recent findings in image diffusion models~\cite{martingarcia2024diffusione2eft}, we demonstrate that strongly conditioned \textit{video} diffusion models can also be fine-tuned as 1-step deterministic models. 
Despite the significantly reduced computational cost, we observe that 1-step models consistently produce more ``accurate'' predictions and outperform multi-step stochastic models in photometric evaluations such as PSNR scores.
However, the 1-step model enforces deterministic output, which can result in blurrier predictions for ambiguous regions with high-frequency details, thus yielding lower perceptual metrics, such as LPIPS. 
For neural forward rendering and relighting tasks, we use multi-step stochastic models to capture more realistic details, though we note that 1-step models can be a competitive choice for error-sensitive tasks and enhancing runtime efficiency.

\begin{figure}[t!]
\centering
\vspace{-5.5mm}
\includegraphics[width=0.99\linewidth]{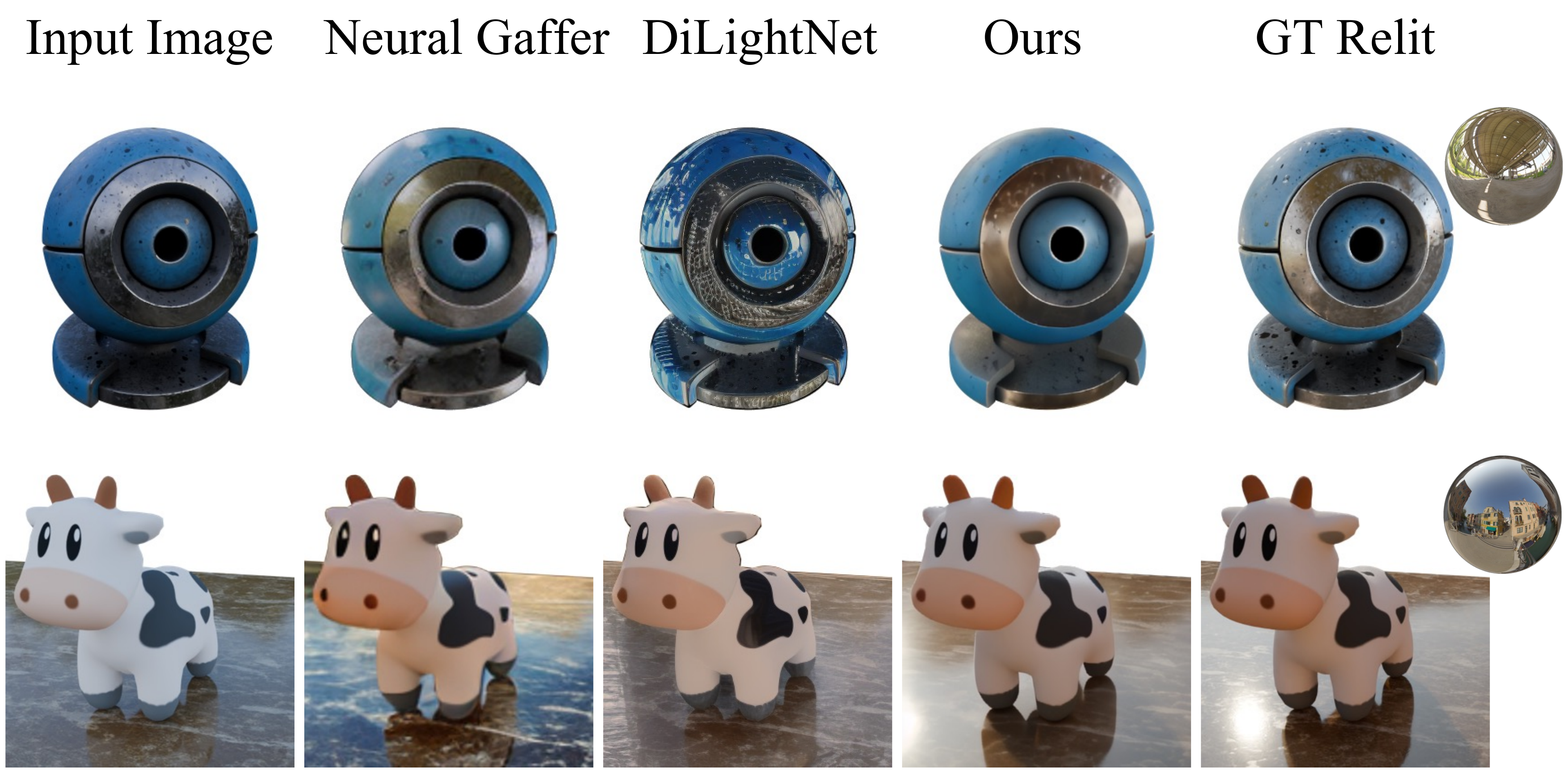}
\vspace{-3mm}
\caption{
\textbf{Qualitative comparison of relighting.} 
Our method produces more accurate specular reflections compared to the baselines.
}
\vspace{-3mm}
\label{fig:relighting_comparison}
\end{figure}

\begin{figure}[t!]
\centering
\includegraphics[width=0.99\linewidth]{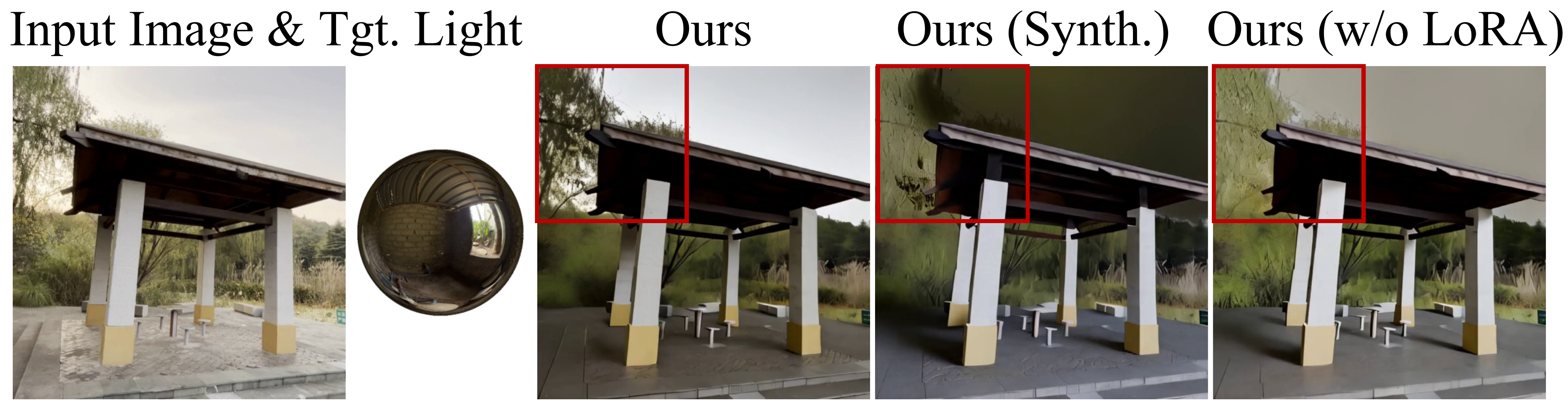}
\vspace{-3mm}
\caption{
\textbf{Qualitative ablation of relighting.} 
Joint training with real-world data and adding LoRA during training significantly improve relighting quality for real-world scenes. 
}
\vspace{-3mm}
\label{fig:relighting_ablation} 
\end{figure} 

\begin{figure}[t!]
\centering
\vspace{-5.5mm}
\includegraphics[width=0.99\linewidth]{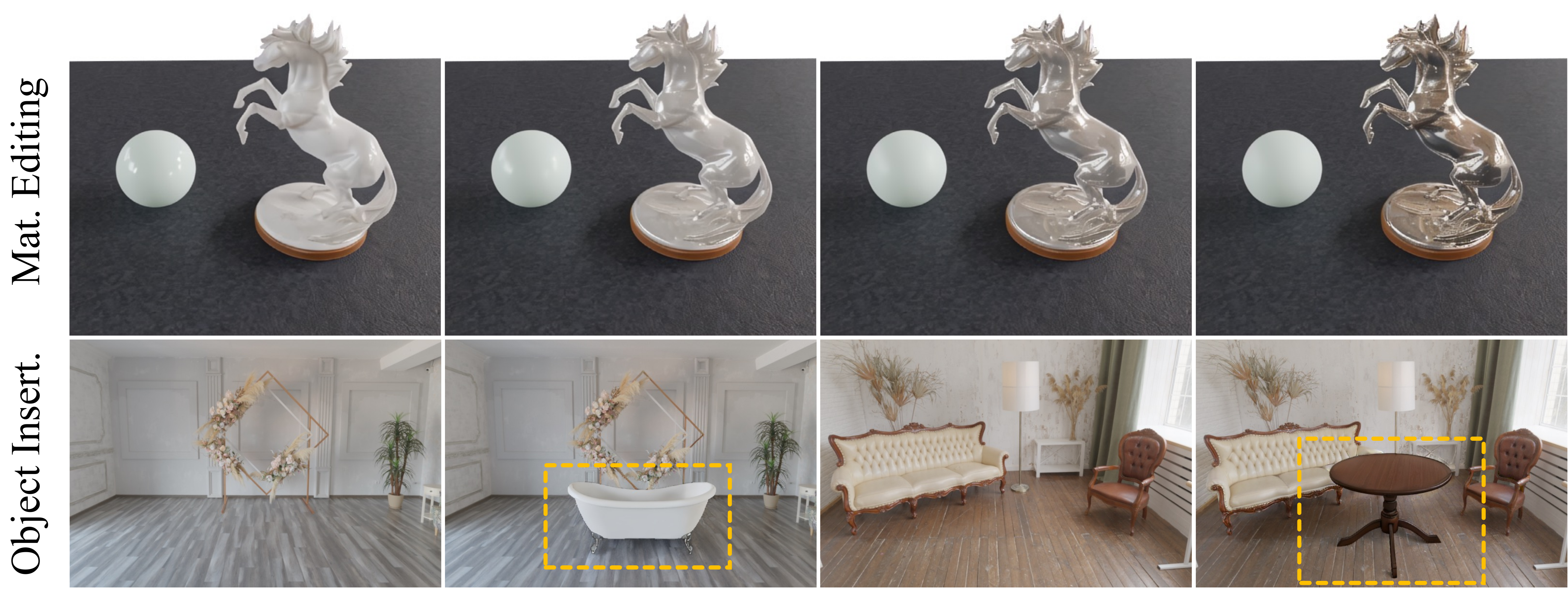}
\vspace{-3mm}
\caption{
\textbf{Image editing applications.} 
Top: Realistic material editing, adjusting the sphere's roughness and the horse's metallic. Bottom: Object insertion of a bathtub and table into scene images. 
}
\vspace{-5mm}
\label{fig:applications} 
\end{figure}

\vspace{-1mm}
\subsection{Evaluation of Relighting}
\label{sec:exp_relighting}
\vspace{-1mm}
In Table~\ref{tab:relighting} and Fig.~\ref{fig:relighting_comparison}, we compare with recent state-of-the-art relighting methods DiLightNet~\cite{zeng2024dilightnet} and Neural Gaffer~\cite{jin2024neural_gaffer}. 
Our method outperforms these baselines, particularly in scenes with complex shadows and inter-reflections.
Overall, it produces high-quality lighting effects and more accurate color and scale.

\parahead{Ablation study.}
We ablate the design choices of synthetic-real joint training in Fig.~\ref{fig:relighting_ablation}. 
While synthetic data provides accurate supervision signals, it is limited to a specific domain and lacks the diversity and complexity found in real-world data. When training exclusively on synthetic data (Ours Synth.), the model struggles with complex structures, such as trees, which are rarely represented in synthetic datasets. 
Since the real-world auto-labels are estimated using inverse rendering models and contain imperfections, we find that incorporating a LoRA~\cite{hu2022lora} during training with real data consistently improves visual quality.

\vspace{-1mm}
\subsection{Applications} 
\label{sec:exp_applications}
\vspace{-1mm}
We show material editing and object insertion applications in Fig.~\ref{fig:applications}.
In the top row, we adjust the sphere's roughness from $0.15$ to $0.6$ and increase the horse's metallic property from $0$ to $1$, achieving photorealistic material edits. 
In the bottom row, we insert a bathtub and table in the G-buffer space of the input image, and use our forward renderer to produce the edited result. The inserted objects blend naturally into the scene, generating realistic reflections and shadows.

%% file: tables/rendering.tex
\begin{table}[t!]
\setlength{\tabcolsep}{2pt}
\centering
\small
\vspace{-2mm}
\resizebox{0.495\textwidth}{!}{
\begin{tabular}{l|ccc|ccc}
\toprule
& \multicolumn{3}{c|}{\textit{SyntheticObjects}} & \multicolumn{3}{c}{\textit{SyntheticScenes}} \\
& PSNR~$\uparrow$ & SSIM~$\uparrow$ & LPIPS~$\downarrow$ & PSNR~$\uparrow$ & SSIM~$\uparrow$ & LPIPS~$\downarrow$ \\
\midrule
SSRT                       				& $\bm{29.4}$ & $\bm{0.951}$ & $\bm{0.037}$ & $\bm{24.8}$ &  $\bm{0.899}$ & $\bm{0.113}$  \\
SplitSum~\cite{karis2013real} 			& $28.7$ & $\bm{0.951}$ & $0.038$ & $23.1$ &  $0.883$ & $0.116$  \\
\midrule
RGB$\leftrightarrow$X~\cite{zeng2024rgb}       		& $25.2$ & $0.896$ & $0.077$ & $18.5$ &  $0.645$ & $0.302$  \\
DiLightNet~\cite{zeng2024dilightnet}   & $26.6$ & $0.914$ & $0.067$ & $20.7$ &  $0.630$ & $0.300$  \\
Ours                       				& $28.3$ & $\bm{0.935}$ & $\bm{0.048}$ & $\bm{26.0}$ &  $\bm{0.780}$ & $\bm{0.201}$  \\
Ours (\texttt{image})           				& $27.4$ & $0.916$ & $0.062$ & $25.4$ &  $0.760$ & $0.215$  \\
Ours (w/o env.~encoder)    				& $27.8$ & $0.927$ & $0.057$ & $25.3$ &  $0.756$ & $0.237$  \\
Ours (+ shading cond.)     			& $\bm{28.7}$ & $0.930$ & $0.056$ & $25.6$ &  $0.761$ & $0.245$  \\
\bottomrule
\end{tabular}
}
\vspace{-3mm}
\caption{
Quantitative evaluation of neural rendering.
}
\vspace{-3mm}
\label{tab:rendering}
\end{table}

%% file: tables/relighting.tex
\begin{table}[t!]
\setlength{\tabcolsep}{2pt}
\centering
\small
\resizebox{0.49\textwidth}{!}{
\begin{tabular}{l|ccc|ccc}
\toprule
& \multicolumn{3}{c|}{\textit{SyntheticObjects}} & \multicolumn{3}{c}{\textit{SyntheticScenes}} \\
& PSNR~$\uparrow$ & SSIM~$\uparrow$ & LPIPS~$\downarrow$ & PSNR~$\uparrow$ & SSIM~$\uparrow$ & LPIPS~$\downarrow$ \\
\midrule
DiLightNet~\cite{zeng2024dilightnet}        & 23.79 & 0.872 & 0.087 & 18.88 & 0.576 & 0.344 \\ %
Neural Gaffer~\cite{jin2024neural_gaffer}   & \emph{26.39} & \emph{0.903} & \emph{0.086} & \emph{20.75} & \emph{0.633} & \emph{0.343} \\
Ours                        				& \textbf{27.50} & \textbf{0.918} & \textbf{0.067} & \textbf{24.63} & \textbf{0.756} & \textbf{0.257} \\
\bottomrule
\end{tabular}
}
\vspace{-2.5mm}
\caption{
Quantitative evaluation of relighting. 
}
\vspace{-5mm}
\label{tab:relighting}
\end{table}

%% file: tables/inverse_rendering_combined.tex
\begin{table*}[thbp]
\setlength{\tabcolsep}{2pt}
\centering
\small
\begin{minipage}{0.65\linewidth}
\centering
\vspace{-6mm}
\resizebox{\textwidth}{!}{ %
\begin{tabular}{l|cccc|c|c|c}
\toprule
& \multicolumn{4}{c|}{Albedo} & Metallic & Roughness & Normals \\
& PSNR~$\uparrow$ & LPIPS~$\downarrow$ & si-PSNR~$\uparrow$ & si-LPIPS~$\downarrow$ & RMSE~$\downarrow$ & RMSE~$\downarrow$ & Angular Error~$\downarrow$ \\
\midrule
RGB$\leftrightarrow$X~\cite{zeng2024rgb} & $14.3$ & $0.323$ & $19.6$ & $0.286$ & $0.441$ & $0.321$ & $23.80^\circ$ \\
Ours  & $\emph{25.0}$ & $\bm{0.205}$ & $26.7$ & $\bm{0.204}$ & $\emph{0.039}$ & $0.078$ & $\emph{5.97}^\circ$ \\
Ours (\texttt{det.}) & $\bm{26.0}$ & $0.219$ & $\bm{27.7}$ & $0.217$ & $\bm{0.028}$ & $\bm{0.060}$ & $\bm{5.85}^\circ$ \\
Ours (\texttt{image}) & $23.4$ & $\emph{0.213}$ & $26.0$ & $\emph{0.209}$ & $0.066$ & $0.098$ & $6.67^\circ$ \\
Ours (\texttt{image}, \texttt{det.}) & $24.8$ & $0.231$ & $\emph{27.2}$ & $0.228$ & $0.043$ & $\emph{0.069}$ & $6.17^\circ$ \\
\bottomrule
\end{tabular}
} %
\vspace{-2.5mm}
\caption{
Quantitative evaluation of inverse rendering on \textit{SyntheticScenes} video dataset. 
\texttt{image}: per-frame inference as image model. \texttt{det.}: 1-step deterministic inference. 
}
\vspace{-3mm}
\label{tab:inverse_rendering}
\end{minipage}%
\hfill
\begin{minipage}{0.325\linewidth}
\centering
\vspace{-6mm}
\resizebox{\textwidth}{!}{ %
\begin{tabular}{l|ccc}
\toprule
 & PSNR~$\uparrow$ & SSIM~$\uparrow$ & LPIPS~$\downarrow$ \\
\midrule
IIW~\cite{bell2014intrinsic} & $9.7$ & $0.62$ & $0.47$ \\
Li~\etal~\cite{li2020inverse} & $12.3$ & $0.68$ & $0.52$ \\
Zhu~\etal~\cite{zhu2022learning} & $15.9$ & $0.78$ & $0.34$ \\
Kocsis~\etal~\cite{kocsis2023iid} & $17.4$ & $\emph{0.80}$ & $0.22$ \\
RGB$\leftrightarrow$X~\cite{zeng2024rgb} & $16.4$ & $0.78$ & $\emph{0.19}$ \\
Ours (\texttt{image}) & $\emph{21.9}$ & $\bm{0.87}$ & $\bm{0.17}$ \\
Ours (\texttt{image}, \texttt{det.}) & $\bm{22.4}$ & $\bm{0.87}$ & $\emph{0.19}$ \\
\bottomrule
\end{tabular}
} %
\vspace{-3mm}
\caption{
Quantitative benchmark of albedo estimation on InteriorVerse dataset~\cite{zhu2022learning}. 
}
\vspace{-3mm}
\label{tab:inverse_rendering_indoor}
\end{minipage}
\end{table*}

%% file: sec/conc.tex
\vspace{-2mm}
\section{Discussion}
\label{sec:conclusion}
\vspace{-1mm}

\ourmodel{} provides a scalable, data-driven approach to inverse and forward rendering, achieving high-quality G-buffer estimation and photorealistic image generation without relying on explicit path tracing or precise 3D scene representations. 
Jointly trained on synthetic and auto-labeled real-world data, \ourmodel{} consistently outperforms state-of-the-art methods.

\parahead{Limitations and future work.} 
Our method is based on Stable Video Diffusion, which operates offline and would benefit from distillation techniques to improve inference speed. 
For editing tasks, the inverse and forward rendering models preserve most of the original content but may introduce slight variations in color or texture. 
Future work could explore task-specific fine-tuning~\cite{kocsis2024lightit} and develop neural intrinsic features to enhance content consistency and handle more complex visual effects.
Additionally, our real-world auto-labeling currently adopts off-the-shelf lighting estimation model~\cite{Phongthawee2023DiffusionLight} which could benefit from better accuracy and robustness. 
With rapid advancements in video diffusion models~\cite{cosmos} toward higher quality and faster inference speeds, we are optimistic that \ourmodel{} will inspire future research in high-quality image synthesis and editing.

%% file: sec/X_suppl.tex
\clearpage
\setcounter{page}{1}
\maketitlesupplementary

\renewcommand{\thefigure}{S\arabic{figure}}
\setcounter{figure}{0}
\renewcommand{\thetable}{S\arabic{table}}
\setcounter{table}{0}
\renewcommand{\thesection}{\Alph{section}}
\setcounter{section}{0}

In the supplementary material, we provide additional implementation details (Sec.~\ref{sec:implementation_details}) and further results and analysis (Sec.~\ref{sec:additional_results}). Please refer to the \textsc{\textcolor{magenta}{accompanying video}} for more qualitative results and comparisons.

\section{Experimental Settings} 
\label{sec:implementation_details}

\parahead{Implementation details.} 
We fine-tune our models based on Stable Video Diffusion\footnote{\url{https://huggingface.co/stabilityai/stable-video-diffusion-img2vid}}~\cite{blattmann2023stable}.

For the \textit{inverse renderer}, we modify the diffusion UNet by expanding four additional channels in the first convolutional layer to include image conditions. We optimize both the diffusion UNet parameters and the domain embedding parameters using a learning rate of $3 \times 10^{-5}$. The training is conducted with a batch size of 256, with a mix of multiple scene attributes. When generating the single-channel depth, metallic, and roughness maps, we average the outputs across the three channels to obtain the final result for each map.

In the \textit{forward renderer}, we expand the first convolutional layer of the diffusion UNet by 20 additional channels to concatenate the additional pixel-aligned G-buffer conditions. Since the depth, metallic, and roughness maps are single-channel properties, we replicate each to create three-channel inputs before passing them into the VAE encoder $\vaeEncoder$. 
The weights of the cross-attention layers are repurposed for lighting conditions, and are reset prior to training. We use a learning rate of $1 \times 10^{-4}$ for optimization. 

Both models are trained using the AdamW optimizer for 20,000 iterations, with mixed-precision (fp16) training at a resolution of 512×512 pixels. 
The training takes around 2 days on 32 A100 GPUs.
We have empirically observed that the video model performs best when trained on video lengths that it will encounter during inference. 
To ensure robust generalization across different frame lengths, we randomly select training video lengths of 1, 4, 8, 16, and 24 frames. This strategy allows the model to adapt effectively to varying video lengths during inference without compromising output quality. 
As a result, the models can also effectively process a single image by treating it as a video with one frame. 
During the training of both models, a 0.1 dropout is applied independently to each condition channel to reduce reliance on individual conditions and potentially enhance robustness. 
During inference, we empirically observe that a small classifier-free guidance (CFG) such as 1.2 enhances the visual quality of the forward rendering model. CFG does not provide noticeable benefit for the inverse rendering model and we do not use it for the inverse rendering model.

\parahead{Data preparation.} 
For synthetic data curation, we begin with the Objaverse~\cite{objaverse} LVIS split, containing 46,207 3D models. 
The 3D assets are filtered based on the following criteria: (i) assets include valid PBR attributes such as roughness and metallic, (ii) assets can be rendered without geometry/texture artifacts. This process yields a final set of 36,500 3D assets. 
We collect 766 HDR panoramas from three sources:
PolyHaven\footnote{\url{polyhaven.com/hdris} (License: CC0)}, 
DoschDesign\footnote{\url{doschdesign.com}(License: \href{https://www.doschdesign.com/information.php?p=2}{link})}, 
and HDRMaps\footnote{\url{hdrmaps.com} (License: Royalty-Free)}.
For PBR textures, we collect 6,300 CC0 textures from multiple sources: 
3D Textures\footnote{\url{https://3dtextures.me/tag/cc0/}}, 
ambientCG\footnote{\url{https://ambientcg.com}}, 
cgbookcase\footnote{\url{https://www.cgbookcase.com/textures}}, 
PolyHaven\footnote{\url{https://polyhaven.com/}}, 
sharetextures\footnote{\url{https://www.sharetextures.com}}, 
and TextureCan\footnote{\url{https://www.texturecan.com}}. 
We remove textures that include only diffuse channels or lack diffuse textures, and manually exclude non-tileable textures, resulting in 4,260 high-quality PBR textures.

In each scene, we place a plane with a randomly selected PBR material, and sample up to three 3D objects, and place them on the plane after randomly rotating, translating, and scaling. We perform collision detection to avoid intersecting objects. 
We also place up to three primitives (cube, sphere, and cylinder) with randomized materials to cover complex lighting effects such as inter-reflections. 
The materials of primitives can be from the aforementioned texture maps or a monolithic material with varying albedo, roughness, and metallic.
A randomly selected HDR environment map illuminates the scene. We also add random horizontal rotation, flipping, and intensity scaling to the environment map.
The rendered videos contain 5 types of motions, 1) 360-degree camera orbits; 2) small-scale regional camera oscillation; 3) 360-degree rotating light with a fixed camera; 4) rotating objects with a fixed camera; and 5) translating objects around the plane.

We render videos of all scenes with corresponding intrinsic images in a custom path tracer based on OptiX~\cite{parker2010optix}, 
with 256~spp, OptiX denoising and AgX tonemapper\footnote{\url{https://github.com/sobotka/AgX}}.
In total, there are 150,000 videos with paired ground-truth G-buffers and environment maps, at 24 frames per video in 512x512 resolution.

\parahead{Baseline configurations.}
DiLightNet~\cite{zeng2024dilightnet} requires a text prompt per example, so we used 
\texttt{meta/llama-3.2-11b-vision-instruct}\footnote{\url{https://www.llama.com/}} to generate a short prompt for each example in \textit{SyntheticObjects} and \textit{SyntheticScenes} based on the first image in each clip and the instruction 
\textit{"What is in this image? Describe the materials. Be concise and produce an answer with a few sentences, no more than 50 words."}

\input{tables/video_metric}

\parahead{Environment map encoder pre-training.} 
As detailed in the main paper, the environment lighting condition in our forward rendering model is encoded through cross-attention between the UNet's spatial latent features and the environment map representation. To provide effective lighting encodings, similar to VAE and CLIP embeddings in diffusion models, we propose pre-training an environment map auto-encoder specifically designed to capture HDR light intensity and orientation.

With both LDR space and log space environment maps ($\envmap_\text{ldr}$  and $\envmap_\text{log}$) as the model input and auto-encoder's reconstruction target, our encoder can retain detailed ambient lighting information while emphasizing high-intensity HDR light spots.
To ensure precise control over light orientation in scene rendering, we introduce a directional encoding map, $\envmap_{\text{dir}}$, where each pixel represents a unit vector corresponding to a light direction in the camera coordinate system. By modifying $\envmap_{\text{dir}}$, the light orientation in the scene can be adjusted accordingly.

\begin{figure}[thbp]
    \centering
    \includegraphics[width=1.0\linewidth]{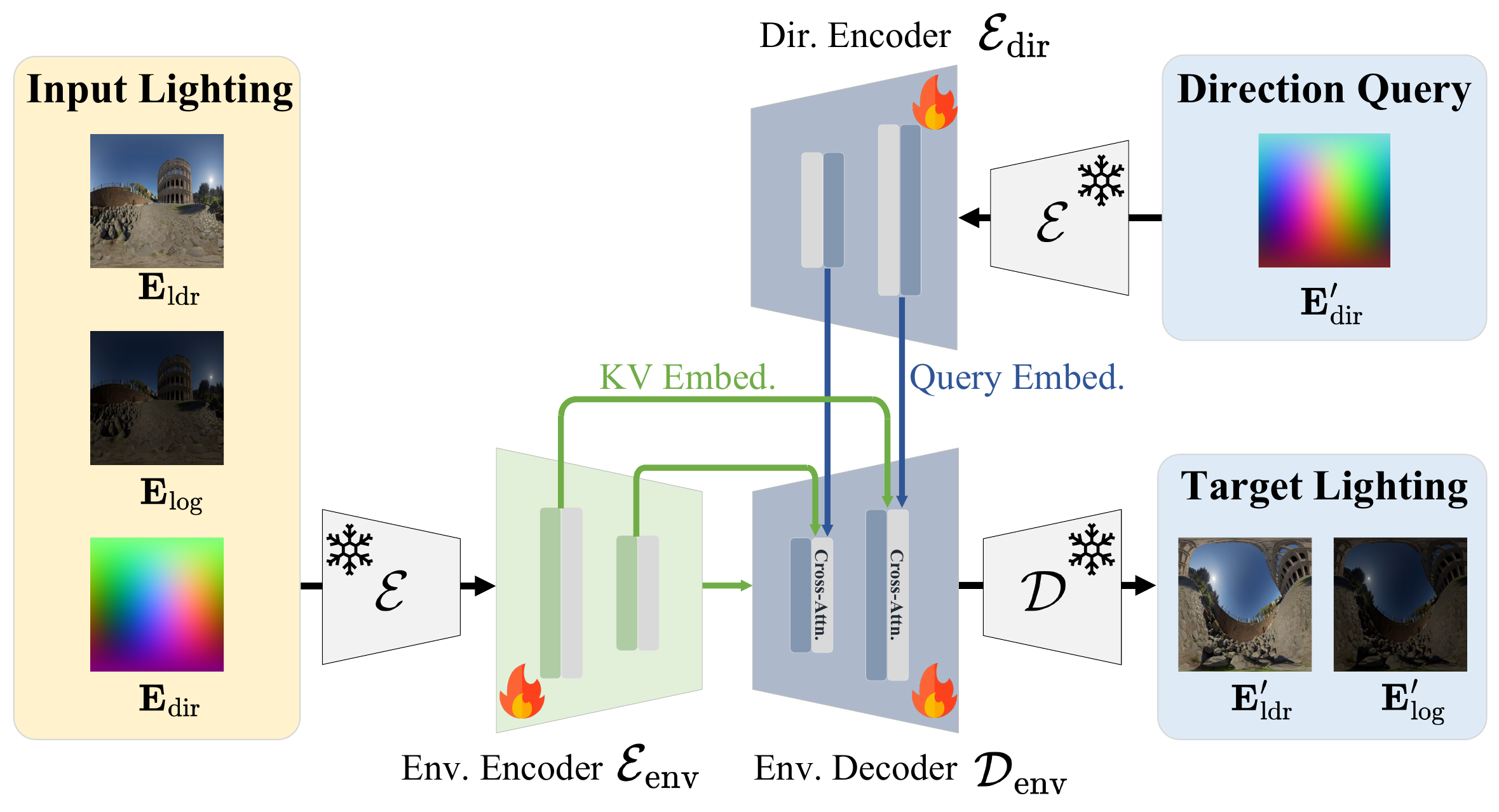}
    \caption{The overview of our environment map auto-encoder training pipeline.}
    \label{fig:env_autoenc}
\end{figure}

The pre-training process aims to produce an environment map encoder $\vaeEncoder_{\text{env}}$, capable of encoding complex directional HDR lighting. For this, we pair $\vaeEncoder_{\text{env}}$ with two auxiliary modules: an environment map decoder $\vaeDecoder_{\text{env}}$ and a direction query encoder $\vaeEncoder_{\text{dir}}$. This forms an auto-encoder training pipeline, as illustrated in Fig. \ref{fig:env_autoenc}.
The encoder $\vaeEncoder_{\text{env}}$ processes concatenated VAE-encoded inputs $\feature_\envmap = ({\vaeEncoder(\envmap_\text{ldr}), \vaeEncoder(\envmap_\text{log}), \vaeEncoder(\envmap_\text{dir})})$, generating $K=4$ levels of multi-resolution features $(\feature_\text{env}^i)_{i=1}^K$. Similarly, $\vaeEncoder_{\text{dir}}$ takes a VAE-encoded directional map $\feature_\mathbf{D} = \vaeEncoder(\envmap'_\text{dir})$, producing features $(\feature_\text{dir}^i)_{i=1}^K$ of the same shape.
The decoder $\vaeDecoder_{\text{env}}$ reconstructs the inputs $\envmap'_\text{ldr}$ and $\envmap'_\text{log}$ using the features $(\feature_\text{env}^i)_{i=1}^K$ and $(\feature_\text{dir}^i)_{i=1}^K$ through cross-attention layers. 
To enhance directional encoding, the training objective involves re-projecting the environment map with random rotations applied to the lighting sphere. This rotation information can be precisely represented by $\envmap^{\prime}_\text{dir}$.
To reconstruct the re-projected environment map, we use the features $( \feature_\text{dir}^i )_{i=1}^K$ encoded from $\envmap^{\prime}_\text{dir}$ as embedding to query the directional HDR lighting encoded in $( \feature_\text{env}^i )_{i=1}^K$ (serving as key-value embedding) through the cross-attention layers in environment map decoder $\vaeDecoder_{\text{env}}$.
The training objective therefore is:
\begin{equation}
    \mathcal{L}_{\text{env}} = \| \feature_{\envmap'} - \vaeDecoder_{\text{env}}(\vaeEncoder_\text{env}(\feature_\envmap), \vaeEncoder_\text{dir}(\feature_\mathbf{D})) \|^2
\end{equation}
where $\feature_{\envmap'} = (\vaeEncoder(\envmap'_\text{ldr}), \vaeEncoder(\envmap'_\text{log}))  \in \mathbb{R}^{\times h_\text{env} \times w_\text{env} \times 8}$.

\begin{figure}
    \centering
    \includegraphics[width=\linewidth]{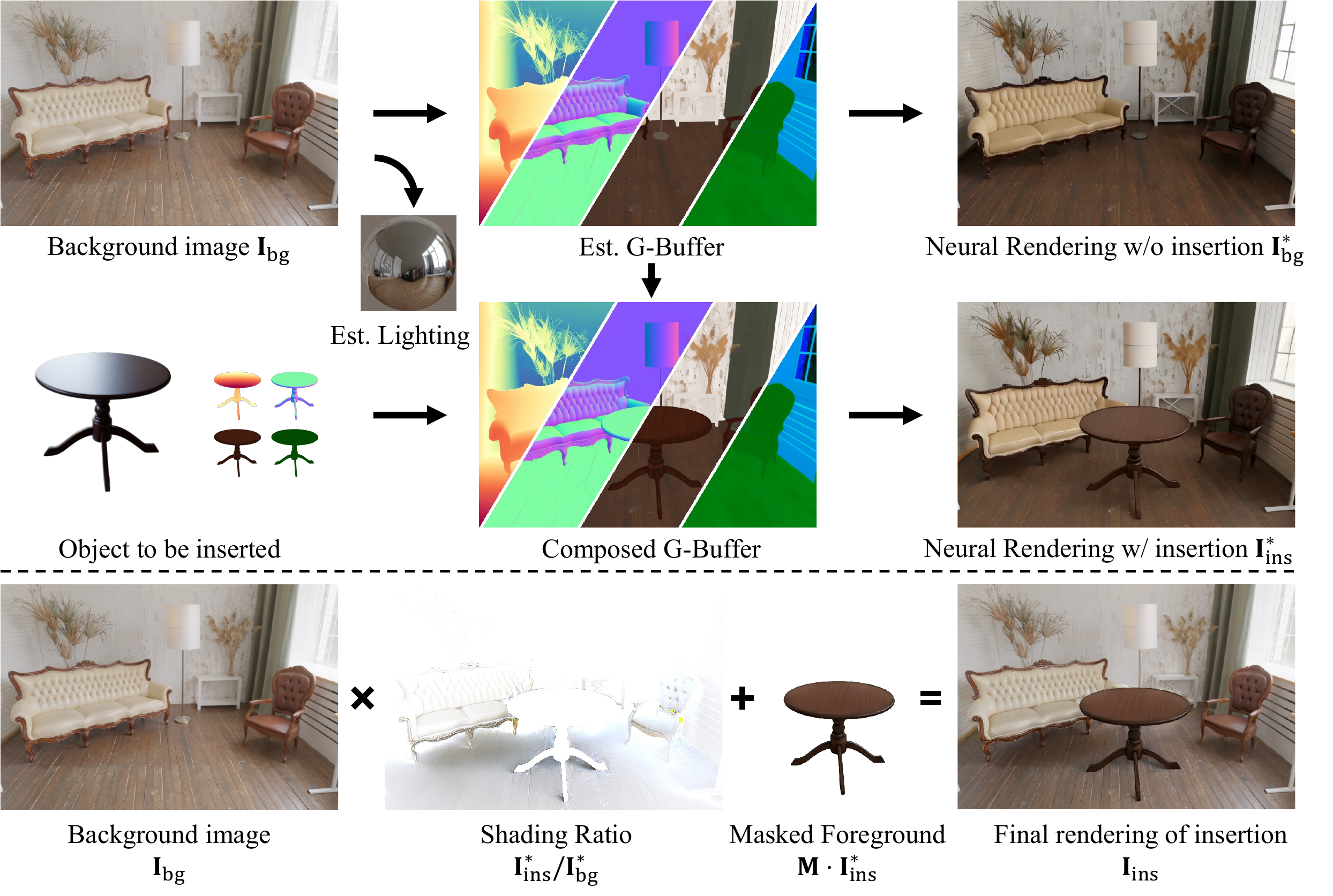}
    \caption{Overview of the object insertion workflow.}
    \label{fig:insetion_workflow}
\end{figure}

\parahead{Object Insertion.}
We provide additional details of object insertion application shown in main paper Fig.~\ref{fig:applications}. 
The objective is to seamlessly insert an object (either 2D or 3D) into a given background image $\image_{\text{bg}}$, ensuring consistent appearance with the background (e.g., aligned lighting effects). 
Our method achieve this task with a combination of the inverse and forward rendering processes, as illustrated in Fig.~\ref{fig:insetion_workflow}.

First, our inverse rendering model estimates the G-buffer of the background image $\image_{\text{bg}}$. 
The G-buffer of the object to be inserted is obtained either through our inverse renderer or directly from a rendering engine. 
Based on the known foreground object mask $\mathbf{M}$, these G-buffers are then blended to create a composite G-buffer. Additionally, we estimate the lighting using an off-the-shelf model \cite{Phongthawee2023DiffusionLight}.

Using the composite G-buffer and estimated lighting, our forward rendering model generates two images: $\image_{\text{ins}}^*$ representing the scene with the inserted object, and $\image_{\text{bg}}^*$, the re-rendering of the original background. 
To minimize unintended changes to the original background image, we follow~\cite{li2020inverse,wang2022neural,liang2024photorealistic} and compute a shading ratio $\bm{\rho} = \image_{\text{ins}}^*/\image_{\text{bg}}^*$ that accounts for the relative shading effects introduced by the inserted object. 

The final edited image $\image_{\text{ins}}$ is computed by multiplying the shading ratio with the original background image $\image_{\text{bg}}$ and compositing the masked foreground object $\mathbf{M} \cdot \image_{\text{ins}}^*$ onto the shaded background:  
\begin{equation}
    \image_{\text{ins}} = (1-\mathbf{M}) \cdot \image_{\text{bg}} \cdot \frac{\image_{\text{ins}}^*}{\image_{\text{bg}}^*} + \mathbf{M}\cdot \image_{\text{ins}}^*
\end{equation}
This process is visualized in Fig.~\ref{fig:insetion_workflow} (bottom).

\section{Additional Results}
\label{sec:additional_results}

\parahead{Runtime cost.} 
Since our models are built on top of Stable Video Diffusion, the inference runtime cost of our models is roughly on the same level as Stable Video Diffusion. 
For a 24-frame video with a resolution of 512x512, the peak GPU memory cost for both models at inference time is around 21 GB. 
the inverse rendering model takes 9.7 seconds to perform 20 denoising steps including VAE encoding and decoding, clocked on one A100 GPU. 
The forward rendering model takes 20.3 seconds to run 20 denoising steps including VAE encoding and decoding. The increased runtime of the forward renderer is due to additional condition signals, which require extra time for encoding. 

Without a separate environment map encoder, Ours (w/o Env. Encoder) completes 20 denoising steps in 19.9 seconds. The runtime overhead introduced by the environment map encoder is negligible.

\begin{figure*}[t!]
\centering
{
\setlength{\tabcolsep}{1pt}
\renewcommand{\arraystretch}{1}
\vspace{-4mm}
\begin{tabular}{cc}
    \includegraphics[width=0.33\linewidth]{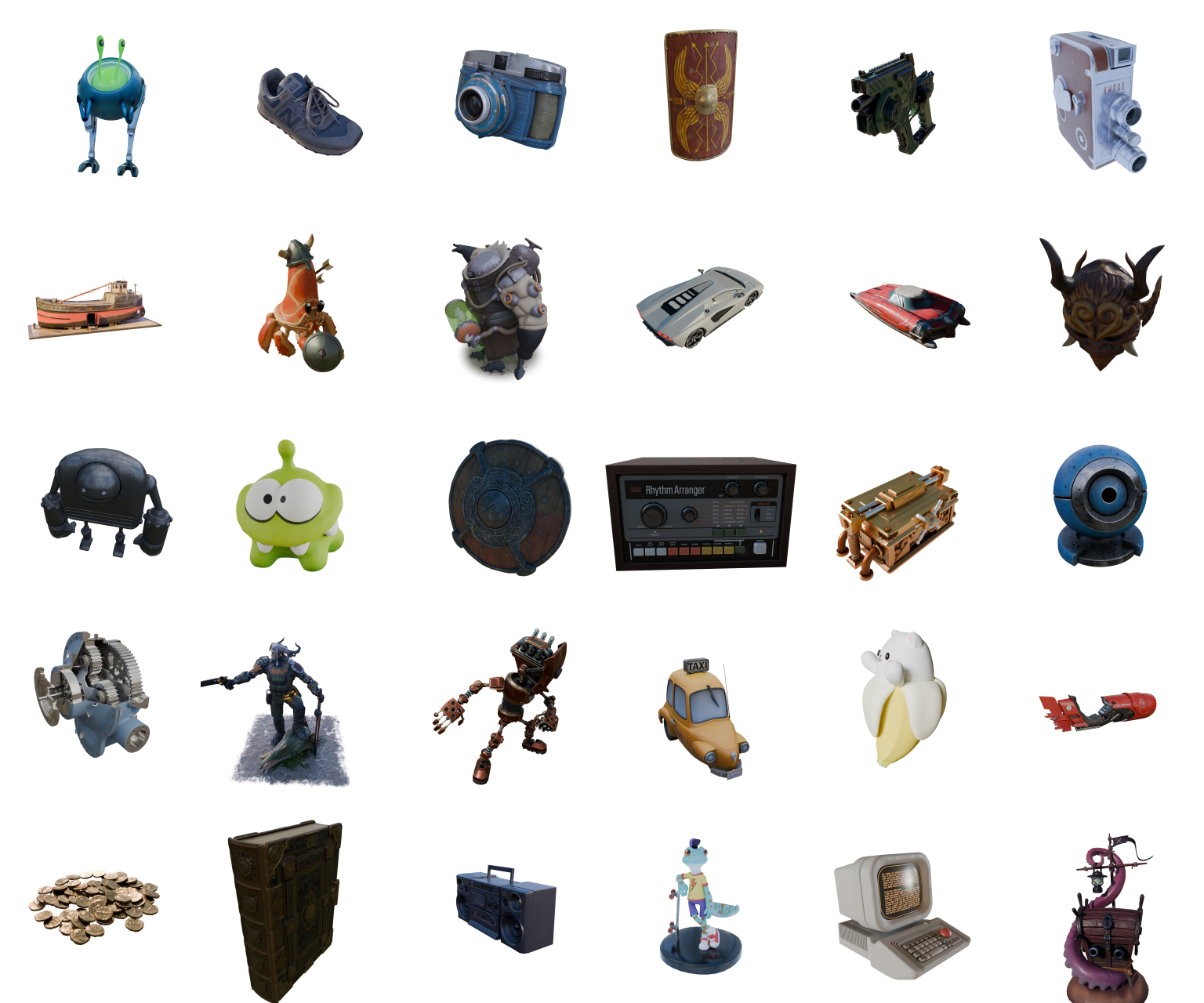} &
    \includegraphics[width=0.66\linewidth]{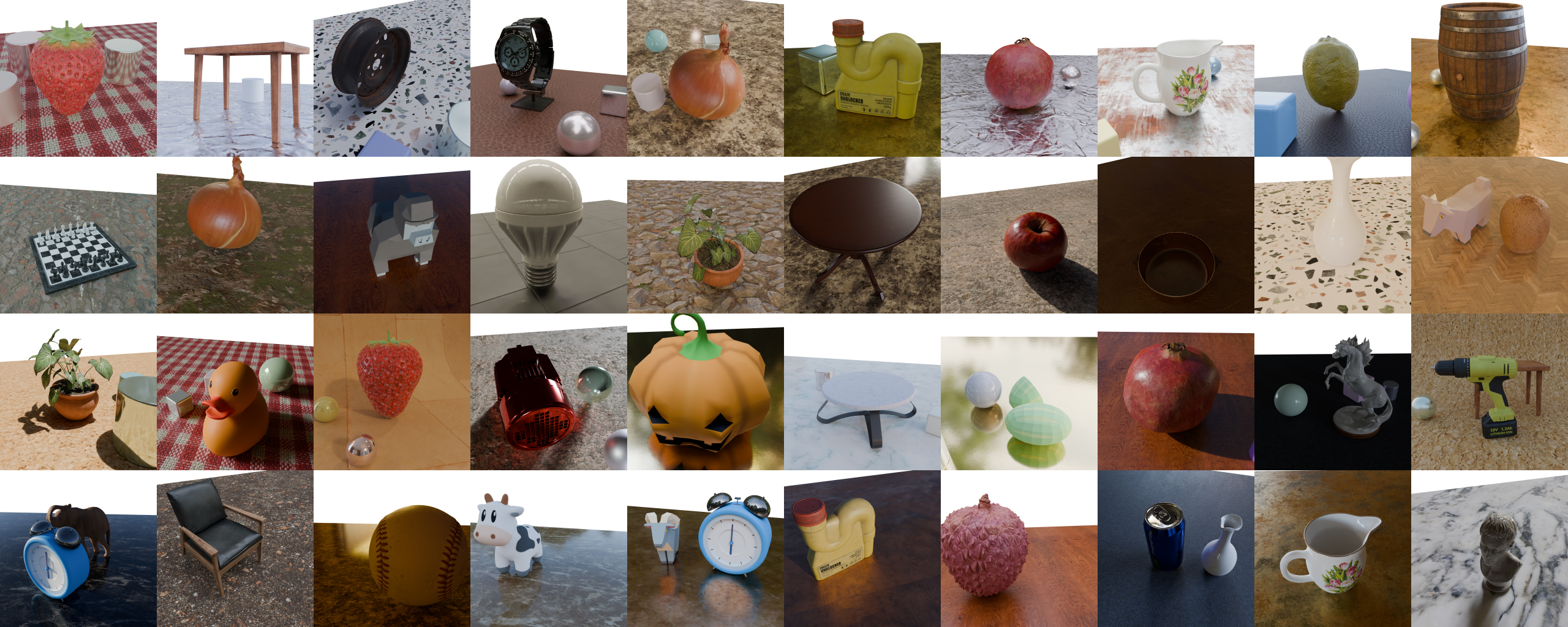} \\
    \textit{SyntheticObjects} & \textit{SyntheticScenes} \\
\end{tabular}
}
\caption{Visualization of the synthetic datasets for quantitative evaluation.}
\label{fig:testset} 
\end{figure*}

\parahead{Temporal consistency.} 
In Table~\ref{tab:video_metric} we report ColorVideoVDP~\cite{mantiuk2024ccvdp} (CVVDP) scores for the relighting comparison (c.f., Table~\ref{tab:relighting} and Fig.~\ref{fig:relighting_comparison} in the main paper). CVVDP predicts the perceptual difference between pairs of videos and accounts for spatial and temporal aspects of vision. We note that our method has the highest CVVDP score for both test sets, which is consistent with visual inspections. Please refer to the supplemental video to assess temporal consistency. In contrast to Neural Gaffer and DiLightNet, which leverage image diffusion models, our approach builds upon \emph{video} diffusion models, which provide considerably improved temporal consistency. For reproducibility, CVVDP was configured according to: 

{\footnotesize
\begin{verbatim}
ColorVideoVDP v0.4.2, 75.4 [pix/deg], Lpeak=200, 
Lblack=0.2, Lrefl=0.3979 [cd/m^2] (standard_4k).
\end{verbatim}
}

\parahead{User study.}
We conducted a user study to evaluate the image perceptual quality of our method. In this study, participants were shown a reference path-traced rendering alongside a pair of renderings: one from our method and one from a baseline (randomly shuffled). They were asked to select which rendering perceptually more closely resembles the reference, considering aspects like lighting, shadows, and reflections. This user study was conducted for both neural rendering and relighting tasks. The evaluation data were sampled from SyntheticScenes and SyntheticObjects (the same datasets used for Table \ref{tab:rendering} and Table \ref{tab:relighting}) (70 scenes).
For each comparison, we collected 9 user selections to determine the preferred rendering by majority voting. The preference percentages for our method compared to baseline approaches are reported across all examples.
Inspired by GPTEval3D~\cite{wu2023gpteval3d}, we repeat this experiment using GPT-4V as perceptual evaluators.
Reported in Table \ref{tab:user_study}, the user study results align with
our findings in the main paper, and indicate a reasonable level
of agreement between human and GPT-4V assessments.

\begin{table}[h]

\input{tables/user_study}

    \caption{
\small\textbf{User study.} 
We report the percentage of images where users preferred Ours over baselines. 
A preference $> 50\%$ indicates Ours outperforming baselines. 
Evaluation follows main paper Table~\ref{tab:rendering}, \ref{tab:relighting} on \textit{SyntheticScenes} and \textit{SyntheticObjects}.
    }
    \label{tab:user_study}
\end{table}

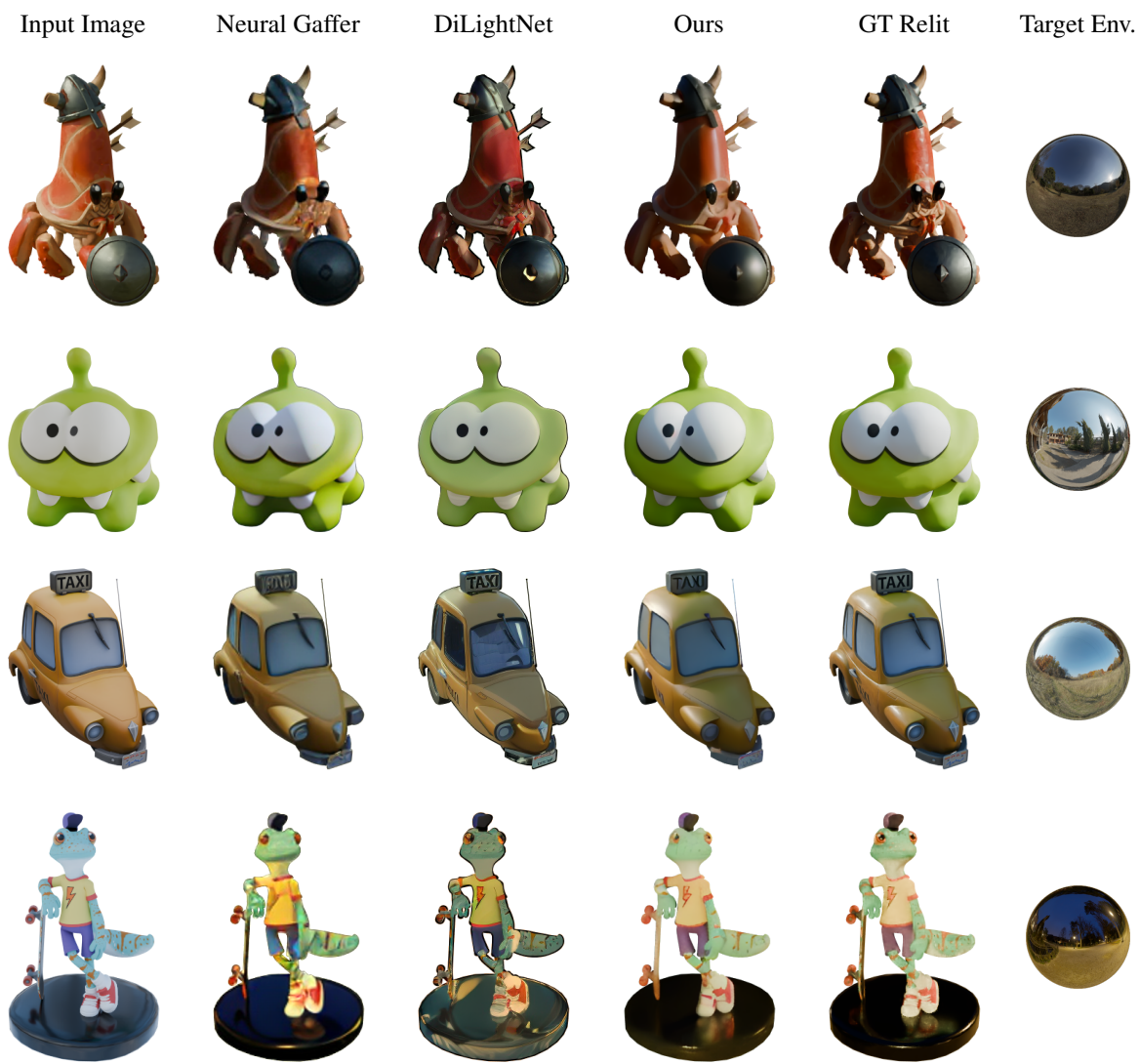
\begin{figure*}
\input{figures/relight_compare}
    \caption{
Additional qualitative comparison of relighting.
Our method produces more accurate specular reflections compared to the baselines.
}
\label{fig:relighting_comparison3}
\end{figure*}

\begin{figure*}[thbp]
\centering
\includegraphics[width=0.99\linewidth]{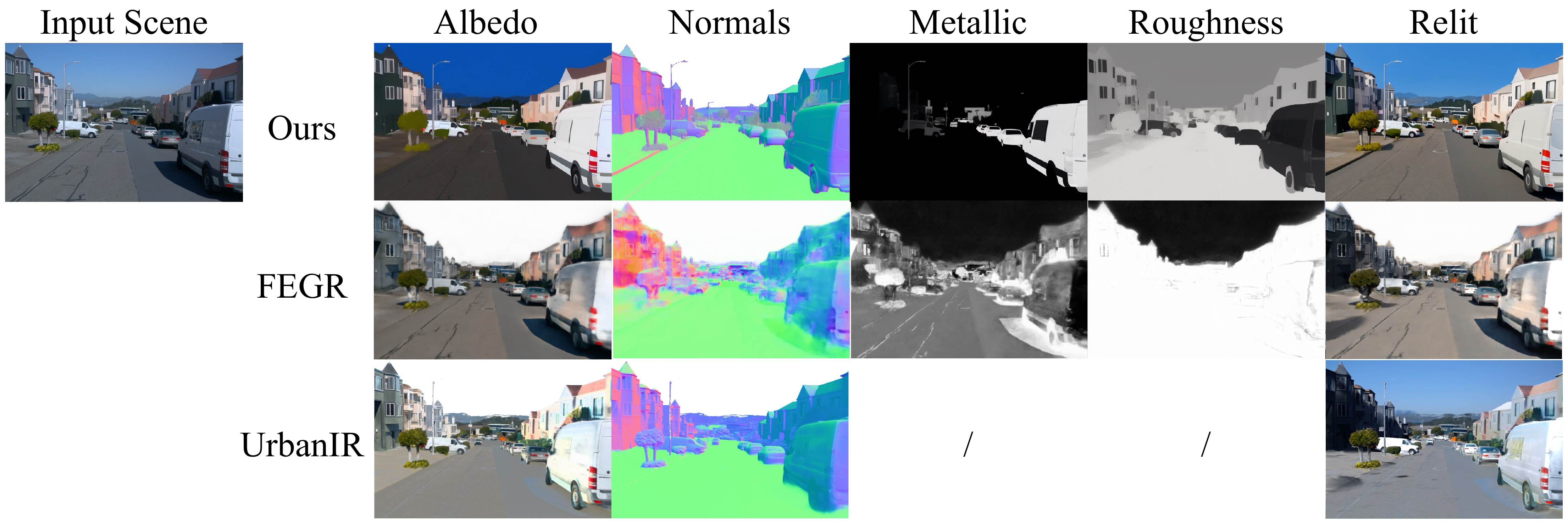}
\includegraphics[width=0.99\linewidth]{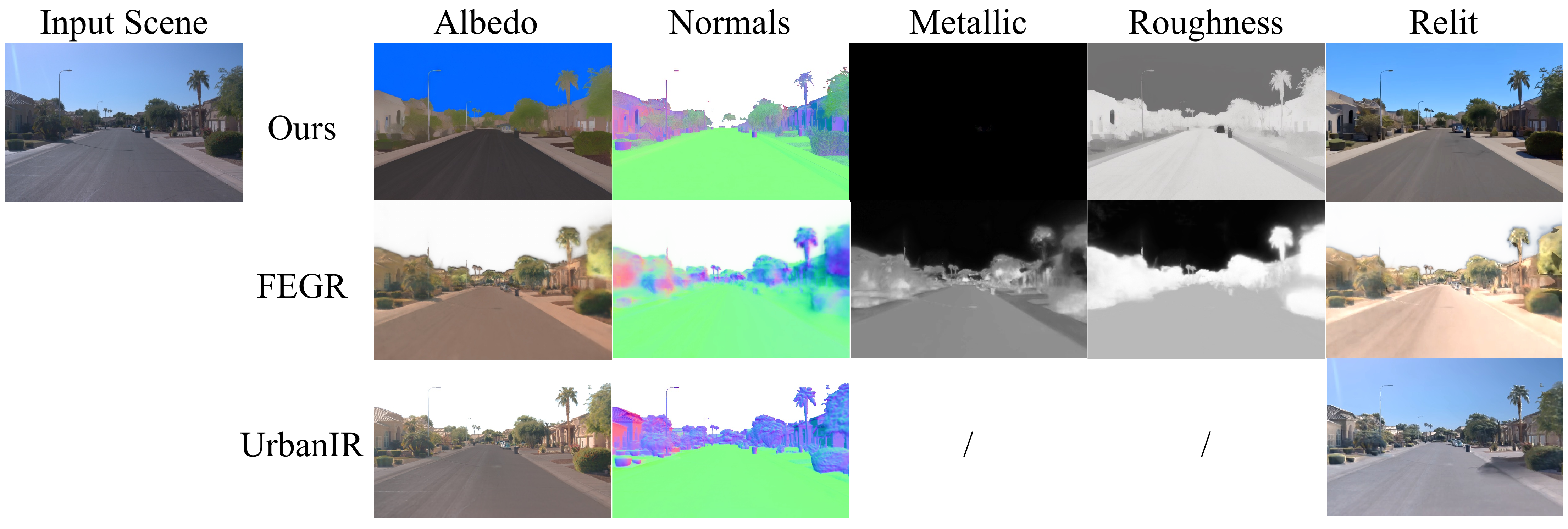}
\vspace{-3mm}
\caption{
Qualitative comparison of inverse rendering and relighting on Waymo dataset with FEGR~\cite{wang2023fegr} and UrbanIR~\cite{lin2023urbanir}. 
}
\vspace{-3mm}
\label{fig:relighting_real}
\end{figure*}

\parahead{Comparison with FEGR~\cite{wang2023fegr} and UrbanIR~\cite{lin2023urbanir}.}
We additionally compare to 3D inverse rendering and relighting approaches FEGR~\cite{wang2023fegr} and UrbanIR~\cite{lin2023urbanir} in Fig.~\ref{fig:relighting_real}. These methods optimize neural 3D representation, then use volume rendering and PBR to produce the final relighting result. 
As the input data is limited to a single illumination condition, they often cannot cleanly remove shadows from the albedo, resulting in shadow artifacts in re-lit results. Additionally, existing scene reconstruction methods struggle to handle highly detailed structures such as trees, and dynamic scenes, which limits their fidelity for PBR path tracing. 
In contrast, our method consistently generates more photorealistic results without relying on explicit 3D geometry constraints. We refer to the accompanying video for animated results. 

%% file: tables/video_metric.tex
\begin{table}[t!]
\setlength{\tabcolsep}{2pt}
\centering
\small
\resizebox{0.4\textwidth}{!}{
\begin{tabular}{l|c|c}
\toprule
CVVDP~$\uparrow$  & {\textit{SyntheticObjects}} & {\textit{SyntheticScenes}} \\
\midrule
DiLightNet~\cite{zeng2024dilightnet}        & 5.44 & 2.99 \\ 
Neural Gaffer~\cite{jin2024neural_gaffer}   & \emph{6.49} & \emph{3.47} \\
Ours                        				& \textbf{6.77} & \textbf{6.40} \\
\bottomrule
\end{tabular}
}
\vspace{-2.5mm}
\caption{
Quantitative evaluation of relighting in terms of ColorVideoVDP. 
ColorVideoVDP reports video quality in the JOD (Just-Objectionable-Difference) units. The highest quality (no difference) is reported as 10 and lower values are reported for distorted content. We compute a JOD value per clip for three novel lighting conditions in each series and report the average over all clips.
}
\vspace{-5mm}
\label{tab:video_metric}
\end{table}

%% file: tables/user_study.tex
\centering
\scriptsize
\setlength{\tabcolsep}{3pt} %
\begin{tabular}{@{} c c *{4}{c} *{2}{c} @{}}
    \toprule
    \multicolumn{2}{c}{} & \multicolumn{4}{c}{\textbf{Neural Rendering}} & \multicolumn{2}{c}{\textbf{Relighting}} \\
    \cmidrule(lr){3-6} \cmidrule(lr){7-8}
    \multicolumn{2}{c}{} & SSRT & SplitSum & RGB$\leftrightarrow$X & DiLightNet & DiLightNet & N.Gaffer \\
    \midrule
    \multirow{2}{*}{\adjustbox{valign=c, rotate=90, margin=0 0pt}{\textit{Scenes}}} 
    & Human    & 72\% & 75\% & 85\% & 85\% & 90\% & 65\% \\
    \addlinespace[2pt]
    & GPT4V    & 40\% & 50\% & 80\% & 85\% & 60\% & 68\% \\
    \midrule
    \multirow{2}{*}{\adjustbox{valign=c, rotate=90, margin=0 3pt}{\textit{Objs}}} 
    & Human    & 37\% & 43\% & 76\% & 83\% & 57\% & 57\% \\
    \addlinespace[2pt]
    & GPT4V    & 57\% & 45\% & 87\% & 54\% & 55\% & 52\% \\
    \bottomrule
\end{tabular}

%% file: figures/relight_compare.tex
\centering
\setlength{\tabcolsep}{5pt}
\begin{tabular}{cccccc}
Input Image & Neural Gaffer & DiLightNet & Ours & GT Relit & Target Env.\\
\raisebox{-0.5\height}{\includegraphics[width=0.14\linewidth]{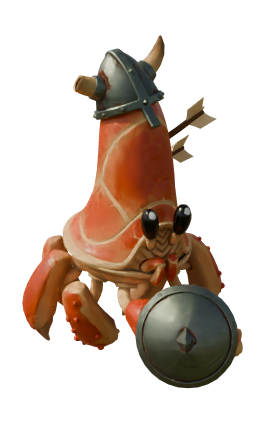}} &
\raisebox{-0.5\height}{\includegraphics[width=0.14\linewidth]{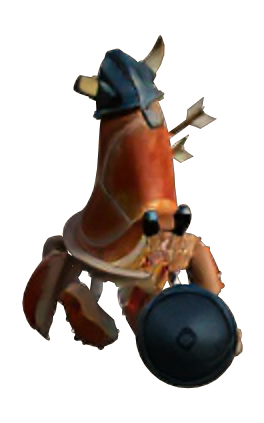}} &
\raisebox{-0.5\height}{\includegraphics[width=0.14\linewidth]{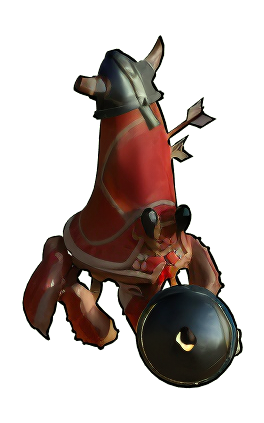}} &
\raisebox{-0.5\height}{\includegraphics[width=0.14\linewidth]{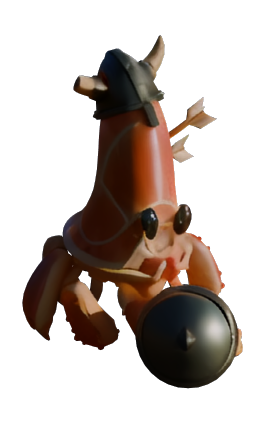}} &
\raisebox{-0.5\height}{\includegraphics[width=0.14\linewidth]{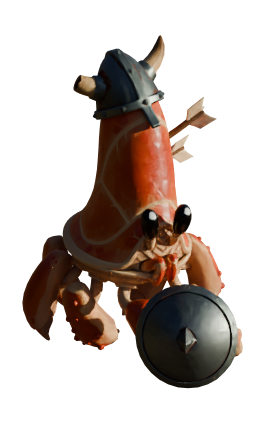}} &
\raisebox{-0.5\height}{\includegraphics[width=0.08\linewidth]{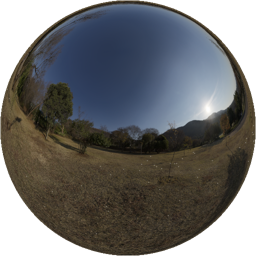}} \\

\raisebox{-0.5\height}{\includegraphics[width=0.14\linewidth]{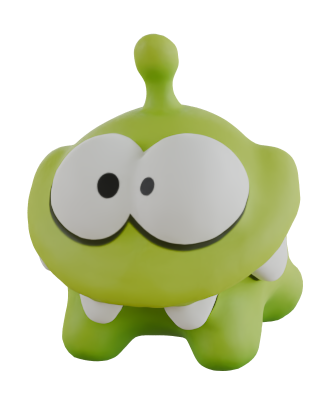}} &
\raisebox{-0.5\height}{\includegraphics[width=0.14\linewidth]{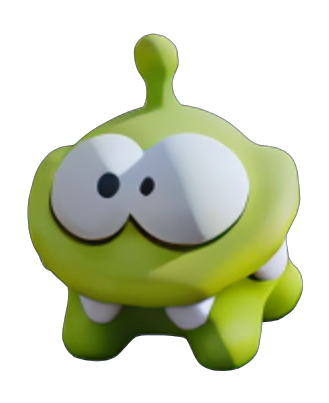}} &
\raisebox{-0.5\height}{\includegraphics[width=0.14\linewidth]{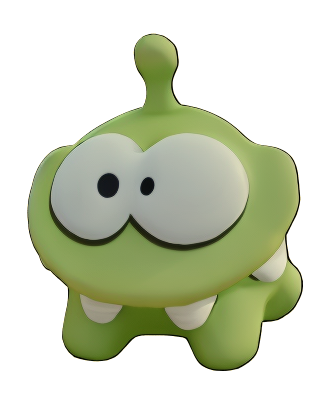}} &
\raisebox{-0.5\height}{\includegraphics[width=0.14\linewidth]{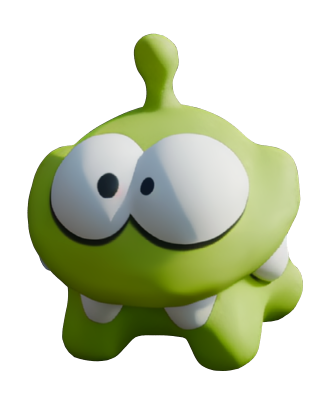}} &
\raisebox{-0.5\height}{\includegraphics[width=0.14\linewidth]{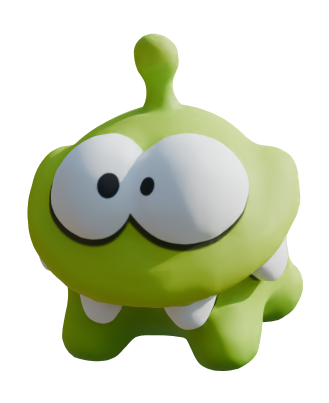}} &
\raisebox{-0.5\height}{\includegraphics[width=0.08\linewidth]{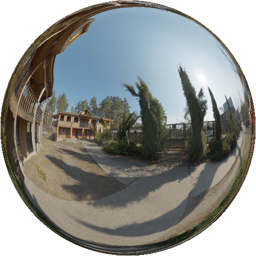}} \\

\raisebox{-0.5\height}{\includegraphics[width=0.14\linewidth]{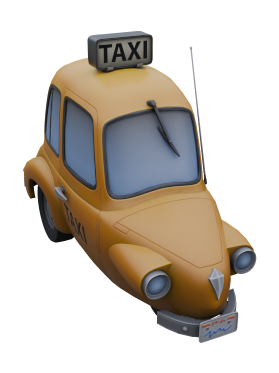}} &
\raisebox{-0.5\height}{\includegraphics[width=0.14\linewidth]{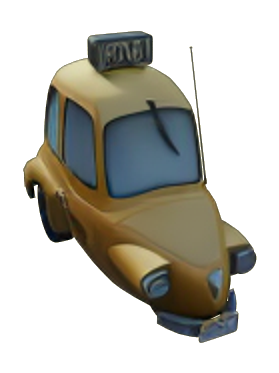}} &
\raisebox{-0.5\height}{\includegraphics[width=0.14\linewidth]{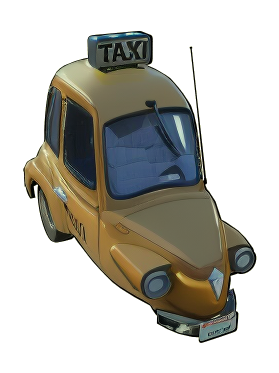}} &
\raisebox{-0.5\height}{\includegraphics[width=0.14\linewidth]{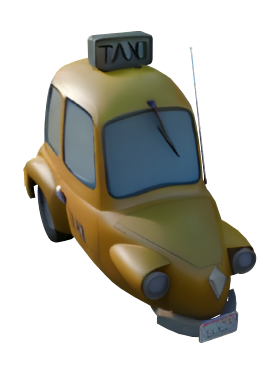}} &
\raisebox{-0.5\height}{\includegraphics[width=0.14\linewidth]{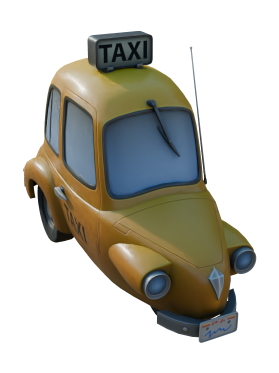}} &
\raisebox{-0.5\height}{\includegraphics[width=0.08\linewidth]{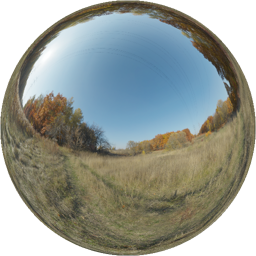}} \\

\raisebox{-0.5\height}{\includegraphics[width=0.14\linewidth]{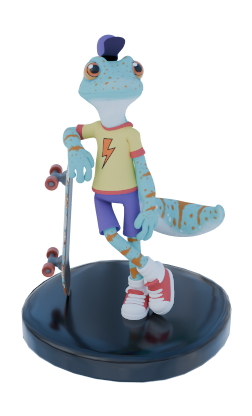}} &
\raisebox{-0.5\height}{\includegraphics[width=0.14\linewidth]{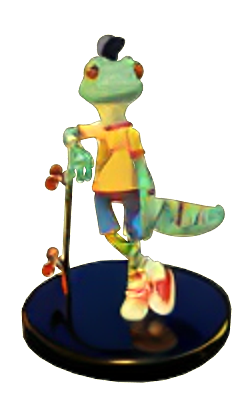}} &
\raisebox{-0.5\height}{\includegraphics[width=0.14\linewidth]{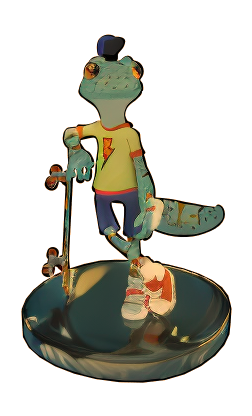}} &
\raisebox{-0.5\height}{\includegraphics[width=0.14\linewidth]{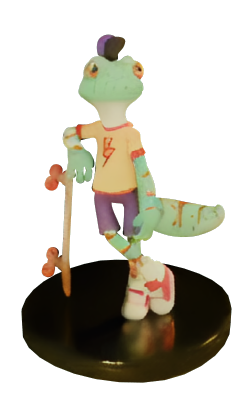}} &
\raisebox{-0.5\height}{\includegraphics[width=0.14\linewidth]{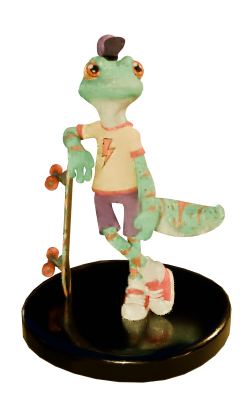}} &
\raisebox{-0.5\height}{\includegraphics[width=0.08\linewidth]{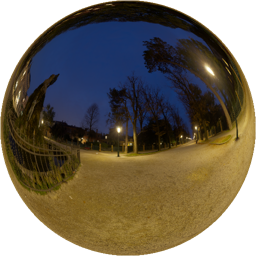}} \\
\end{tabular}